\documentclass{clv3}

\usepackage{booktabs}
\usepackage{linguex}
\usepackage{amsmath,amssymb,mathtools,bm}
\usepackage{tikz}
\usepackage{enumitem}
\usepackage{multirow}

\usepackage{csquotes}

\usepackage{hyperref}
\usepackage{xcolor}
\definecolor{darkblue}{rgb}{0, 0, 0.5}
\hypersetup{colorlinks=true,citecolor=darkblue, linkcolor=darkblue, urlcolor=darkblue}

\usepackage{cleveref}
\newcommand{\lmparams}{\bm{\theta}}
\newcommand{\lmprob}{p_{\lmparams}}
\newcommand{\s}{w_1w_2\ldots{}w_n}

\crefname{ExNo}{}{}
\crefname{SubExNo}{}{}

\creflabelformat{SubExNo}{(#2#1#3)}
\creflabelformat{ExNo}{(#2#1#3)}
\crefrangelabelformat{SubExNo}{(#3#1#4--#5\crefstripprefix{#1}{#2}#6)}
\crefrangelabelformat{ExNo}{(#3#1#4)--(#5#2#6)}

\Crefname{itemize}{}{}

\begin{document}
\issue{$\infty$}{$\infty$}{2024}

\dochead{}

\runningtitle{Generating event descriptions}

\runningauthor{Cao, Holt, Chan, Richter, Glass, \& White}

\pageonefooter{All experimental materials, data, and code used in this paper are available at \url{https://github.com/superMereo/generating-event-descriptions}.}

\title{Generating event descriptions under syntactic and semantic constraints}

\author{Angela Cao}
\affil{University of Rochester}

\author{Faye Holt}
\affil{Georgia Institute of Technology}

\author{Jonas Chan}
\affil{Georgia Institute of Technology}

\author{Stephanie Richter}
\affil{University of Rochester}

\author{Lelia Glass}
\affil{Georgia Institute of Technology}

\author{Aaron Steven White}
\affil{University of Rochester}

\maketitle

\begin{abstract}
With the goal of supporting scalable lexical semantic annotation, analysis, and theorizing, we conduct a comprehensive evaluation of different methods for generating event descriptions under both syntactic constraints---e.g. desired clause structure---and semantic constraints---e.g. desired verb sense. 
We compare three different methods---(i) manual generation by experts; (ii) sampling from a corpus annotated for syntactic and semantic information; and (iii) sampling from a language model (LM) conditioned on syntactic and semantic information---along three dimensions of the generated event descriptions: (a) naturalness, (b) typicality, and (c) distinctiveness.  
We find that all methods reliably produce natural, typical, and distinctive event descriptions, but that manual generation continues to produce event descriptions that are more natural, typical, and distinctive than the automated generation methods. 
We conclude that the automated methods we consider produce event descriptions of sufficient quality for use in downstream annotation and analysis insofar as the methods used for this annotation and analysis are robust to a small amount of degradation in the resulting event descriptions.
\end{abstract}

\section{Introduction}
\label{sec:introduction}



The development and evaluation of an empirically robust lexical semantic generalization requires at least two kinds of data: 
\begin{enumerate}[label=(\roman*),itemsep=-1pt]
    \item a representative sample of lexical items that fall under that generalization;
    \item for each such lexical item, 
    \begin{enumerate}[label=(\alph*),itemsep=-1pt]
        \item a representative sample of the linguistic expressions that each lexical item can (and perhaps, cannot) be used in, along with their context of use;
        \item information about the inferences supported by the lexical item in conjunction with its context.
    \end{enumerate}
\end{enumerate}
In this paper, we investigate three ways of satisfying requirement (iia) as a means for supporting downstream lexical semantic annotation, analysis, and theorizing: 
(1) traditional manual generation of linguistic expressions by experts; 
(2) sampling linguistic expressions from an annotated corpus; 
and (3) sampling linguistic expressions from a language model (LM). 

Our proximal aim is to assess the quality of the linguistic expressions that are produced by each method when that method is required to enforce specific syntactic and semantic constraints on those lexical items.
This proximal aim serves our overarching aim: 
to assess whether automated generation methods, such as methods (2) and (3), are of sufficiently high quality to use in downstream, large-scale lexical semantic annotation, analysis, and theorizing in the absence of \textit{post hoc} human correction.
Insofar as they are safe to use, these automated methods may provide a means of implementing scalable semantic annotation that allows analysts and theoreticians to better target lexical semantic properties of interest than existing ``semantic bleaching'' methods \citep{white_computational_2016,white_role_2018,white_frequency_2020,an_lexical_2020,moon_source_2020,kane_intensional_2022}. 

We focus on sampling sentences in English under heavy syntactic and semantic constraints.
This focus on English allows us to satisfy requirement (i) by drawing on existing verb lexicons, such as VerbNet \citep{kipper-schuler_verbnet:_2005} and PropBank \citep{palmer_proposition_2005}, which classify verbs in terms of their senses and subcategorization properties (among other things).
And given that English has substantial resources in terms of both annotated corpora and state-of-the-art language models, this focus additionally allows us to interpret our results as a sort of upper bound on what one can expect in terms of the quality of automatically generated sentences.

In three experiments, we compare each method along three dimensions: (a) how natural the sentences it produces are; (b) how typical the (kinds of) events described by the sentences are; and (c) how distinctive the sentences are. 
High quality in terms of naturalness is desirable in the sense that unnatural sentences cannot be guaranteed to satisfy particular syntactic and semantic constraints specified by the analyst. 
For instance, \Cref{ex:intro-runner-run-it} is unnatural because of its ill-formed object. 

\ex. The runner ran the it.\label{ex:intro-runner-run-it}

And while one might be able to recover that the type of event described in \Cref{ex:intro-runner-run-it} is similar to the one described in \Cref{ex:intro-runner-run-marathon}---e.g. on the basis of the well-formed subject---it is generally more desirable to not require an annotator to perform that sort of recovery, since different annotators may do so in different ways---potentially introducing noise into the measure of interest.

\ex. The runner ran the marathon.\label{ex:intro-runner-run-marathon}

High quality in terms of typicality is similarly desirable in that atypical event descriptions cannot be guaranteed to satisfy particular semantic constraints specified by the analyst.
For instance, in contrast to \Cref{ex:intro-runner-run-it}, the object of \Cref{ex:intro-table-run-marathon} is syntactically well-formed; but \Cref{ex:intro-table-run-marathon} is clearly atypical of the sorts of running described in \Cref{ex:intro-runner-run-marathon}.

\ex. The table ran the marathon.\label{ex:intro-table-run-marathon}

Depending on the sort of annotation of interest, this atypicality may not matter, but we take it that it is generally more desirable to present annotators with typical examples of a lexical item \citep[cf.][]{reisinger_semantic_2015,white_universal_2016,white_universal_2020}.

Finally, high quality in terms of distinctivenss is desirable in the sense that the example is well-targeted for the semantic property of interest. For instance, \Cref{ex:intro-someone-run-something} can be reasonably thought of as both a natural and a typical description of the same event as \Cref{ex:intro-runner-run-marathon}; however, \Cref{ex:intro-someone-run-something} could just as easily be used to describe the same event as \Cref{ex:intro-ceo-run-company}. 

\ex. Someone ran something.\label{ex:intro-someone-run-something}

\ex. The CEO ran the company.\label{ex:intro-ceo-run-company}

Thus, \Cref{ex:intro-someone-run-something} is intuitively less distinctive---i.e. more general---than either \Cref{ex:intro-ceo-run-company} or \Cref{ex:intro-runner-run-marathon}. 
As for naturalness and typicality, such generality may or may not be a problem, depending on the property of interest.
But there are certainly cases where it could present an issue. For example, if one were interested in assessing whether \textit{run} is telic in a transitive, interpreting \Cref{ex:intro-someone-run-something} as involving the same sense as \Cref{ex:intro-table-run-marathon}, which is a telic description, gives different results than interpreting \Cref{ex:intro-someone-run-something} as involving the same sense as \Cref{ex:intro-ceo-run-company}, which is an atelic description.

We discuss the benefits and drawbacks of each generation method in \Cref{sec:three-approaches-to-generating-examples} before turning, in \Cref{sec:implementing-the-three-approaches}, to our specific implementation of each method to generate materials for our three experiments. 
In \Crefrange{sec:experiment-1-naturalness}{sec:experiment-3-distinctiveness}, we describe our naturalness, typicality, and distinctiveness experiments, in which we find that the automated methods reliably produce natural, typical, and distinctive sentences, but that manual generation continues to produce sentences that are more natural, typical, and distinctive than these automated generation methods. 
We conclude, in \Cref{sec:conclusion}, that automated methods are of sufficient quality to use as assistive technologies for scaling research in lexical semantics insofar as downstream annotation and analysis tolerates a small amount of degradation in the linguistic expressions to be annotated and analyzed. 

\section{Three approaches to generating examples}
\label{sec:three-approaches-to-generating-examples}

We consider each method---manual generation in \Cref{sec:manual-generation-by-experts}, sampling from a corpus in \Cref{sec:sampling-from-a-corpus}, and sampling from a language model in \Cref{sec:sampling-from-a-language-model}---alongside its upsides and downsides (summarised in \Cref{tab:gen-summary}). For the sake of concreteness, we couch our discussion in terms of generating sentences, since that is our focus in this paper; but our discussion applies equally to generating smaller or larger linguistic expressions.

\begin{table}[t]
\centering
\caption{Preliminary summary of sampling methods for linguistic stimuli.}
\label{tab:gen-summary}
\begin{tabular}{rccc}
\toprule
                & \textbf{Quality} & \textbf{Effort} & \textbf{Efficiency} \\
                \midrule
\textbf{Manual} & High             & High             & Low                 \\
\textbf{Corpus} & ?                & Low              & Medium              \\
\textbf{LLM}    & ?                & Low              & High                \\
\bottomrule
\end{tabular}

\end{table}

\subsection{Manual generation by experts}
\label{sec:manual-generation-by-experts}

On the one hand, manually generated sentences can be carefully controlled for a variety of factors that may influence the inferences supported by such a sentence, such as its clausal and nominal structure, as well as morphology.
This fine-grained control over such factors in turn supports apples-to-apples comparisons among lexical items.

For example, in a context where we are interested in understanding the properties of the verb \emph{hit} and related predicates, one may aim to generate sentences like those in \Cref{ex:manual} as exemplars for further investigation \citep[sentences from][p. 127]{fillmore_grammar_1970}. 

\ex. \label{ex:manual} 
     \a. John hit the tree (with a rock).
     \b. A rock hit the tree.

These sentences are carefully controlled in the sense that: 
(i) they are monoclausal and lack temporal or modal adjuncts adjuncts, negative polarity items, etc. that could influence the inferences supported by the sentences or the acceptability; 
(ii) they contain only common names or simple singular (in)definite noun phrases headed by nouns that stereotypically enter into events of hitting; 
and (iii) they contain a verb in a simple tense that does not introduce modality, which could substantially alter the inferences that the sentences support.
They can furthermore be interpreted with minimal context, in part because they describe relatively stereotypical situations.  

In return for this careful control, it is generally expensive and inefficient to generate sentences---especially in cases where an expert is required to ensure that constraints such as those described above are satisfied. 
This expense and inefficiency presents a challenge in cases where one is interested in developing broad-coverage lexical semantic generalizations on the basis of a large sample, since it may be infeasible to manually generate a sufficiently large sample for downstream annotation in a reasonable amount of time.  

\subsection{Sampling from a corpus}
\label{sec:sampling-from-a-corpus}

A less expensive, more efficient approach to sampling linguistic expressions---common in usage-based approaches---is to draw them from a corpus \citep[see][for a review]{katz_semantics_2019}. 
For example, to study \textit{hit} using this method, we might sample sentences like those in \Cref{ex:corpus} from a corpus like the Pushshift Reddit dataset \citep{baumgartner_pushshift_2020}.

\ex. \label{ex:corpus}  
\a. This is something that really hit me hard when I started working at a school.\label{ex:corpus-school} 
\b. When coal regulations hit here, there were bread lines. 

The main challenge in using such data is that the resulting sentences are not controlled for factors that may influence the inferences supported by an expression. 
As exemplified in \Cref{ex:corpus}, corpus data are often multi-clausal and grammatically complex---perhaps in ways orthogonal to one's purpose. 
They furthermore often require a rich context for their interpretation, and it is not always clear how much extrasentential context one must retain to ensure that the sentence is interpretable---or if such context is not provided, how the reader infers a likely context on the basis of the sentence itself and what effects that has on measures of interest. 
For example, in \Cref{ex:corpus-school}, one has to infer antecedents for several pronouns and interpret \textit{hit} as abstract rather than physical, which may not be the sense of interest.

Additional control can be imposed on sentences sampled from a corpus by using syntactic and semantic annotations to filter the sample. 
For instance, if one is interested in sampling only examples of \textit{hit} in a transitive clause under its sense of physical contact, syntactic annotations might be combined with word sense annotations to find only sentences that satisfy the relevant constraints. 

But this approach gives rise to a further challenge: as additional constraints are imposed on a sample, sentences satisfying those constraints become fewer and further between, and thus the size of the corpus must grow to ensure a sample of consistent size. 
This growth may need to be potentially quite substantial given that lexical sentences are power law distributed (\citealt{zipf_psychobiology_1936, zipf_human_1949}, see \citealt{piantadosi_zipfs_2014} for a review), meaning that truly massive corpora may be required for ensuring that sufficient sample sizes for lower frequency lexical sentences are achieved.\footnote{
    One way to deal with this issue is to simply ignore low-frequency lexical items. 
    But one does this at their own peril. 
    See \citealt{white_role_2018} for evidence that focusing too heavily on high-frequency items has led to incorrect lexical semantic generalizations.
} 
Thus, while sampling from a corpus is certainly less expensive than manual generation, its efficiency is dependent on the desired level of control over the syntax and semantics of the sampled sentences.

One way to deal with the challenge posed by sparsity is to combine corpus sampling, constrained by syntactic and semantic information, with post-processing. For instance, \Cref{ex:corpus-edit1-raw} is an example from the Pushshift Reddit dataset that we extracted by looking for instances of \textit{hit} in a transitive construction---a relatively light constraint retaining many matches. In a case where we want an example of \textit{hit} satisfying the heavier constraints we can impose on manually generated sentences, we can then use an automatic procedure that makes use of the sentence's syntactic parse to yield an item such as \Cref{ex:corpus-edit1-postedit}.    

\ex. \label{ex:corpus-edit1}
\a.  \textbf{Raw:} I feel like all the New Years resolutioners must hit the gym on 2 January as it was this morning.\label{ex:corpus-edit1-raw}
\b. \textbf{Post-edited:} The resolutioners hit the gym.\label{ex:corpus-edit1-postedit} 

Insofar as the annotations used to enforce the constraints are of high quality, this procedure will definitionally produce sentences that satisfy the relevant constraints, but it is not known whether these sentences are valid in the same way one can ensure that manually generated sentences are. 
A main aim of this paper is to assess the quality of the sentences that result from a method of this form.

\subsection{Sampling from a language model}
\label{sec:sampling-from-a-language-model}

An alternative way to deal with the challenge posed by sparsity is to work with a compressed form of a corpus, such as a language model (LM). 
An LM describes the distribution of strings in a corpus as a probability distribution $\lmprob$ with parameters $\lmparams$. 
These parameters are estimated (roughly) by obtaining those for which $\lmprob$ assigns maximum likelihood to the strings in the corpus.\footnote{
    Many contemporary language models are estimated using methods that go beyond maximum likelihood estimation. 
    A simple case of this is the incorporation of regularization terms. 
    Another, very common, more complex case involves \textit{preference tuning}, e.g., using reinforcement learning from human feedback \citep[][i.a.]{christiano_deep_2017,stiennon_learning_2020,ouyang_training_2022}.
}

The upshot of describing the distribution of strings in a corpus as a probability distribution is that, insofar as one knows how to sample from $\lmprob$, the LM may be used to (noisily) sample from the corpus that it was estimated on. 
Assuming that $\lmprob$ furthermore assigns non-zero probability to strings that satisfy the desired syntactic and semantic constraints, one can sample sentences that satisfy those constraints from:
\[\bar{p}_{\langle\bm{\psi}, \lmparams\rangle}(\s \mid \text{\sf{}constraints}) \propto p'_{\bm{\psi}}(\text{\sf{}constraints} \mid \s) \times \lmprob(\s)\]
One generic way to implement sampling from $\bar{p}_{\langle\bm{\psi}, \lmparams\rangle}$ is rejection sampling:
(i) sample a string $\s$ from $\lmprob$; 
and (ii) accept that sample with probability $p'_{\bm{\psi}}(\text{\sf{}constraints} \mid \s)$.
A major issue with this approach is that is can be extremely inefficient---potentially far more inefficient than simply sampling from the corpus itself---since the vast majority of samples drawn from $\lmprob$ will not satisfy the constraints. 

At least two approaches can be used to mitigate this inefficiency:
(a) sampling from a language model conditioned on a ``prompt'' that encodes the constraints \citep[][i.a.]{radford_language_2018,brown_language_2020};
and (b) modifying the language model's probabilities so that strings that satisfy the constraints have higher probability \citep[\textit{constrained sampling};][\textit{i.a.}]{holtzman_learning_2018,dathathri_plug_2020,yang_fudge_2021}.\footnote{
    The literatures on both prompting and constrained sampling are vast, and there are many ways to implement both. We focus here on relatively simple variants that do not require us to change anything about the underlying language model---or indeed, train a new system at all---since we believe this setting is the most realistic for most linguists. 
}
Both approaches take advantage of the fact that $\lmprob$ is decomposable as:
\[\lmprob(\s) \equiv q_{\lmparams}(w_1) \times q_{\lmparams}(w_2 \mid w_1) \times 
 \ldots{} \times q_{\lmparams}(w_N \mid w_1 \ldots w_{N-1})\]
where $q_{\lmparams}$ is a probability distribution on lexical items conditioned on strings.
A string can then be sampled by incrementally sampling $w_i$ from $q_{\lmparams}$ conditioned on $w_1 \ldots w_{i-1}$.

\subsubsection{Sampling from a language model conditioned on a prompt}

The LM prompting approach encodes the constraints to be satisfied as natural language strings. For instance, to enforce the semantic constraint that a sentence sampled from an LM contain the verb \textit{hit} in the sense of \textit{strike}, one might sample from $q_{\lmparams}$ conditioned on \Cref{ex:llm-strike-prompt}, where \Cref{ex:llm-strike-prompt} is simply viewed as a string that the LM itself could have generated.

\ex. The following is an example of a sentence that contains the verb ``hit'' in its sense meaning ``strike'':\label{ex:llm-strike-prompt}

The sentences in \Cref{ex:llm-strike} are sampled from the LM \texttt{llama-2-13B} \citep{touvron_llama_2023} conditioned on this prompt.

\ex. \label{ex:llm-strike} 
\a. The baseball hit the bat.\label{ex:llm-strike-baseball}
\b. He hit the ball out of the park.\label{ex:llm-strike-ball}
\c. Joe hit me with his car.\label{ex:llm-strike-car}
\d. John hit me on the head.\label{ex:llm-strike-head}

Anecdotally, this approach works surprisingly well for simple semantic constraints, such as the one expressed in \Cref{ex:llm-strike-prompt}. 
However, it can be difficult to know exactly how to specify syntactic and morphological constraints in a general way.
For example, perhaps we want all of the arguments to be definite noun phrases (NPs)---e.g. because we want the sentences to be maximally evocative of the relevant sense and also make more sense out of context. Or, perhaps we do not want the prepositional phrases (PPs) that appear in \Cref{ex:llm-strike-ball}--\Cref{ex:llm-strike-head}---e.g. because the structure in \Cref{ex:llm-strike-ball}--\Cref{ex:llm-strike-head} could produce importantly different inferences than the simple transitive.

\subsubsection{Constrained sampling}

Enforcing such constraints is where the constrained sampling approach shines.
In constrained sampling, one specifies constraints on the strings to be sampled as an auxiliary probability distribution on those strings.\footnote{
    Constrained sampling is closely related to \textit{constrained decoding}, which is a widely-used approach in structured prediction that attempts to obtain an output with maximal probability (or more generally, score), rather than sampling outputs from a constrained distribution. See \citealt{smith_linguistic_2011} for a general overview of structured prediction for linguistic data, and \citealt{deutsch_general-purpose_2019,shin_constrained_2021} (and references therein) on constrained decoding. A main reason to use constrained sampling, rather than constrained decoding, is that the resulting strings tend to be higher quality \citep{holtzman_curious_2019}.
} 
These constraints can in principle be arbitrarily complex---e.g. requiring access to an entire string \citep[][i.a.]{holtzman_learning_2018}. 
But as discussed above, this complexity can make sampling difficult---or at the very least, require the estimation of additional models, which we aim to avoid.

To mitigate this difficulty, we consider only constraints that are themselves decomposable into a conditional distribution $c_{\bm{\gamma}}$ with parameters $\bm{\gamma}$.
These constraints can then be combined with the incremental sampling procedure by sampling from a constrained distribution $\bar{q}_{\langle\lmparams, \bm{\gamma}\rangle}$, rather than $q_{\lmparams}$ itself.
\[\bar{q}_{\langle\lmparams, \bm{\gamma}\rangle}(w_i \mid w_1 \ldots{} w_{i-1}) \propto q_{\lmparams}(w_i \mid w_1 \ldots{} w_{i-1}) \times c_{\bm{\gamma}}(w_i \mid w_1 \ldots{} w_{i-1})\]
The distribution $c_{\bm{\gamma}}$ can take a variety of forms. In the case where one wants to enforce syntactic constraints, a natural way to state them is in terms of a probabilistic context free grammar (PCFG), where $\bm{\gamma}$ gives rule probabilities. Then, $c_{\bm{\gamma}}(w_i \mid w_1 \ldots{} w_{i-1})$ can be computed using, e.g., an Earley parser \citep{earley_efficient_1970,stolcke_efficient_1995}.\footnote{
    This use of probabilistic Earley parsers is well-known in the psycholinguistics literature for its use in evaluating information-theoretic models of sentence processing \citep[see][\textit{et seq}]{hale_probabilistic_2001,levy_expectation-based_2008}. 
    No stock should be placed in our suggestion of a PCFG over a more expressive formalism, such as a probabilistic linear context free rewriting system \citep[][]{kato_stochastic_2006,kallmeyer_data-driven_2013}.
    The main criterion that must be satisfied is that a conditional distribution over the next word be computable. 
}

For instance, if one wants to enforce the syntactic constraints discussed in \Cref{sec:manual-generation-by-experts}---that all arguments be definite NPs and that the sorts of prepositional phrases (PPs) that appear in \Cref{ex:llm-strike-ball}--\Cref{ex:llm-strike-head} be excluded---one might define $c_{\bm{\gamma}}$ in terms of \Cref{ex:grammar}.

\ex. \texttt{S -> NP VP}\\\texttt{NP -> D N}\\\texttt{VP -> V NP}\\\texttt{V -> hit}\\\texttt{D -> the}\\\texttt{N  -> .+}\label{ex:grammar}

In this grammar, the \texttt{S}, \texttt{NP}, \texttt{VP}, \texttt{V}, and \texttt{D} rules would all necessarily have probability 1---and thus they impose hard constraints on the samples---and \texttt{N  -> .+} is to be interpreted such that \texttt{N} can be rewritten equiprobably as any non-empty string. The latter would be too permissive if this grammar were intended to model transitive clauses headed by \textit{hit}, but because the language model provides information about words that are likely to come after \textit{the}, further constraint is unnecessary.

The result of defining $c_{\bm{\gamma}}$ in terms of \Cref{ex:grammar} is to enforce that \Next. 

\ex. 
\a. $\bar{q}_{\langle\lmparams, \bm{\gamma}\rangle}(\text{the}) = 1$\\\phantom{------}because $c_{\bm{\gamma}}(\text{the}) = 1$
\b. $\bar{q}_{\langle\lmparams, \bm{\gamma}\rangle}(w \mid \text{the}) = q_{\lmparams}(w \mid \text{the})$\\\phantom{------}because $c_{\bm{\gamma}}(w \mid \text{the}) \propto 1$ for all $w$
\c. $\bar{q}_{\langle\lmparams, \bm{\gamma}\rangle}(\text{hit} \mid \text{the}, w) = 1$\\\phantom{------}because $c_{\bm{\gamma}}(\text{hit} \mid \text{the}, w) = 1$ for all $w$
\d. $\bar{q}_{\langle\lmparams, \bm{\gamma}\rangle}(\text{the} \mid \text{the}, w, \text{hit}) = 1$\\\phantom{------}because $c_{\bm{\gamma}}(\text{the} \mid \text{the}, w, \text{hit}) = 1$ for all $w$
\e. $\bar{q}_{\langle\lmparams, \bm{\gamma}\rangle}(w' \mid \text{the}, w, \text{hit}, \text{the}) = q_{\lmparams}(w' \mid \text{the}, w, \text{hit}, \text{the})$\\\phantom{------}because $c_{\bm{\gamma}}(w' \mid \text{the}, w, \text{hit}, \text{the}) \propto 1$ for all $w, w'$

The main downside of this approach to constrained sampling is that the distribution over earlier words---e.g. the noun in the subject of \textit{hit}---cannot be constrained by later words---e.g. \textit{hit} itself and the noun coming after \textit{hit}.\footnote{
    An alternative route to constrained sampling that does not have this downside is to use a masked language model, which is trained to predict the identity of elements $w_{i_1}, \ldots, w_{i_k}$ in a string $\s$ given all elements of the string besides $w_{i_1}, \ldots, w_{i_k}$ \citep{devlin_bert_2019}. For instance, one could in principle sample $w_2$ from $p(\,\cdot \mid \text{the}, \text{--}, \text{hit}, \text{the}, \text{--})$ and then $w_5$ from $p(\,\cdot \mid \text{the}, w_2, \text{hit}, \text{the}, \text{--})$.
    We do not take this route for two reasons: (i) anecdotally, using a masked language model---specifically, RoBERTa \citep{liu_roberta_2019}---produced worse samples than an autoregressive language model like \texttt{llama-2-13b}; and likely relatedly, (ii) the literature has largely abandoned masked language modeling in recent years, meaning that the largest---and therefore, generally most performant---language models are autoregressive. Thus, we deemed it more practical to focus on autoregressive models. 
}
We mitigate this downside in two ways: (i) by combining prompting and constrained sampling as a means to provide information about the verb that will be generated before that verb's subject is generated; and (ii) by generating multiple samples that are then reranked by their probability under the language model (discussed in \Cref{sec:implementing-the-three-approaches}). 

\subsubsection{Combining prompting and constrained sampling}
\label{sec:combining-prompting-and-constrained-sampling}

Prompting and constrained sampling can be combined quite easily. 
However, it remains an open question whether---under the sorts of heavy constraints lexical semanticists work---the sentences sampled using this method are valid in the same way one can ensure that manually generated sentences are.
For instance, does prompting a language model with a gloss of a verb's sense actually produce good examples of that verb in that sense? 
\Cref{ex:llm-reach} shows four sentences sampled from \texttt{llama-2-13b} conditioned on the prompt in \Cref{ex:llm-reach-prompt}. While \Cref{ex:llm-reach-road} is a good example of the \textit{reach, encounter} sense of \textit{hit}, \Crefrange{ex:llm-reach-ball}{ex:llm-reach-head-wall} are better examples of the \textit{strike} sense.\footnote{
    Both of these sense glosses come from PropBank's manually constructed \href{https://github.com/propbank/propbank-frames/blob/main/frames/hit.xml}{frame file for \textit{hit}}: \texttt{hit.01} (\textit{strike}) and \texttt{hit.02} (\textit{reach, encounter}).
}  

\ex. The following is an example of a sentence that contains the verb ``hit'' in its sense meaning ``reach, encounter'':\label{ex:llm-reach-prompt}

\ex. \label{ex:llm-reach} 
\a. We hit the road at dawn.\label{ex:llm-reach-road}
\b. The ball hit him squarely on his forehead.\label{ex:llm-reach-ball}
\c. He hit his head on the door frame.\label{ex:llm-reach-head-door}
\d. She hit her head on the wall.\label{ex:llm-reach-head-wall}

Our aim in the remainder of the paper is to formally assess these questions in a series of judgment studies focused on the naturalness, typicality, and distinctiveness of expressions generated under the set of methods described above.  


\section{Implementing the three approaches}
\label{sec:implementing-the-three-approaches}

To compare the three approaches discussed in \Cref{sec:three-approaches-to-generating-examples} along our three dimensions of interest, we use each method to generate sentences constructed from a well-controlled set of verbs. All sentences are generated under the constraints specified in \Cref{ex:constraints}.

\ex. \label{ex:constraints}
\a. The sentence must be a monoclausal transitive of the form \texttt{NP V NP}.
\b. The verb must be in its simple past tense form.
\b. The subject and object of the transitive must be definite noun phrases of the form \texttt{the N}.
\b. Given a verb and a sense of that verb, the interpretation of the sentence must be compatible with that sense. Assuming that the sense is itself compatible with a transitive, this compatibility is solely determined by the nouns in the subject and object.

We describe how we select verbs for inclusion in our sentences in \Cref{sec:choosing-verbs} and ensure that all constraints can be satisfied for those verbs. In \Cref{sec:manually-generating-examples,sec:implementing-sampling-from-a-corpus,sec:generating-examples-from-a-language-model}, we describe how we implement the generation approaches laid out in \Cref{sec:three-approaches-to-generating-examples}. 

\subsection{Choosing verbs}
\label{sec:choosing-verbs}

Verbs were chosen with reference to their senses in the PropBank lexicon \citep{palmer_proposition_2005} and their classification according to \citealt{levin_english_1993}, as encoded in the VerbNet lexicon \citep{kipper-schuler_verbnet:_2005}, which groups verbs by their syntactic behavior---e.g. VerbNet's \texttt{hit-18.1} class, which includes \textit{beat}, \textit{kick}, \textit{smash}, and so on.\footnote{
    VerbNet was originally conceived as a digitization of \citeauthor{levin_english_1993}'s classification. VerbNet has expanded beyond this original classification \citep{kipper_extending_2006}, but for the subset of classes that correspond to ones in the original classification, the numerical identifier associated with the class corresponds to the (sub)section of \citealt{levin_english_1993} in which that class is discussed.   
}  
Using these resources, we selected 32 triplets of verbs (96 verbs) reflecting the criteria in \Cref{ex:verb-criteria}.\footnote{
    We take the number of senses that PropBank associates with a particular verb as a rough proxy for how polysemous it is. See \citealt{hovy_ontonotes_2006} on data-driven approaches to validating PropBank senses using interannotator agreement. 
} \Cref{tab:verb-triplets} gives examples of three such triplets.

\begin{table}[t]
\centering
\caption{Example verb triplets used in our study.}
\label{tab:verb-triplets}
\begin{tabular}{cccc}
      \toprule
      \textbf{VerbNet Class} & \textbf{Polysemous} & \textbf{Monosemous} & \textbf{Calibration}  \\
      \midrule
       \texttt{hit-18.1}    & hit   & kick   & smash \\
       \texttt{judgment-33} & abuse & insult & mock \\
       \texttt{clear-10.3}  & clear & clean  & drain \\
        \bottomrule
\end{tabular}
\end{table}

\ex. \label{ex:verb-criteria}
\a. All verbs in the triplet share a Levin class in their transitive form.
\b. One verb in the triplet is polysemous in its transitive form according to PropBank, with 3-5 PropBank senses in its transitive form.
\b. One verb is monosemous in its transitive form according to PropBank, with 1 PropBank sense in its transitive form.
\b. One verb, which we use for our manually generated and calibration sentences, has 1 or 2 senses in its transitive form according PropBank.
\b. All verbs in the triplet were manually judged as very similar in meaning, on some sense of the polysemous members.

We choose both polysemous (e.g. \textit{hit}) and monosemous (e.g. \textit{kick}) verbs from each class in order to explore how well each automated method is able to generate sense-distinct sentences in cases where a verb has multiple senses. We select a third \textit{calibration verb} (e.g. \textit{smash}) in order to produce manually generated sentences that we use for two purposes: (i) to assess the extent to which experts can reliably generate sentences that vary with respect to naturalness, typicality, and distinctiveness; and (ii) to calibrate participants judgments in each of the three tasks described in \Crefrange{sec:experiment-1-naturalness}{sec:experiment-3-distinctiveness}. We describe both uses in more detail in \Crefrange{sec:experiment-1-naturalness}{sec:experiment-3-distinctiveness}.



\subsection{Manual generation by experts} 
\label{sec:manually-generating-examples}

We wrote sentences by hand for our 32 calibration verbs.
These sentences were partitioned into two \textit{calibration sets}.
The \textit{core calibration set} was used across all experiments; the \textit{extended calibration set} was only used in the distinctiveness experiments for reasons discussed below.

\subsubsection{Core calibration set}

For our first set of calibration sentences, we wrote four sentences for each verb: 
one that we deemed to be a \textit{natural} description of a \textit{typical} situation; 
one that we deemed to be a \textit{natural} description of an \textit{atypical} situation; 
one that we deemed to be an \textit{unnatural} description of a \textit{typical} situation; 
and one that we deemed to be an \textit{unnatural} description of an \textit{atypical} situation.
This procedure resulted in 124 manually generated sentences.

\paragraph{Manually generating natural, typical sentences}

For each calibration verb, we first chose a subject and object deemed \textit{natural} and \textit{typical} when used with the verb in question. 
The distinction we draw between these two dimensions is that natural sentences follow all structural rules of English, while typical sentences use content nouns that play an expected role in the situation described by the verb---e.g. cooks commonly \textit{smash} potatoes as in \Cref{ex:natural_typical}.

\ex. The cook smashed the potatoes.\label{ex:natural_typical}

\paragraph{Manually generating natural, atypical sentences}

Next, we changed either the subject or object of the natural and typical sentence to another noun phrase such that the result is \textit{natural} but \textit{atypical}. The atypical sentences use content nouns that play a surprising role in that situation---e.g. it is unusual for a strawberry to \textit{smash} a potato, as in \Cref{ex:natural_atypical}.

\ex. The strawberry smashed the potatoes.\label{ex:natural_atypical}

\paragraph{Manually generating unnatural, typical sentences}

To generate the \textit{unnatural} but \textit{typical} sentence, we took the natural and typical sentence and changed the noun phrase that was unchanged for the natural but atypical sentence. 
Unnatural sentences contain syntactic violations---e.g. combining the pronoun \textit{them} with a definite determiner, as in \Cref{ex:unnatural_typical}. 

\ex. The cook smashed the them.\label{ex:unnatural_typical}

\paragraph{Manually generating unnatural, atypical sentences}

Finally, the \textit{unnatural} and \textit{atypical} sentence takes the two new noun phrases introduced by \Cref{ex:natural_atypical} and \Cref{ex:unnatural_typical} and replaces those in \Cref{ex:natural_typical}---exemplified in \Cref{ex:unnatural_atypical}.

\ex. The strawberry smashed the them.\label{ex:unnatural_atypical}

There are an equal number of sentences in which the subject is replaced for the natural and atypical sentences as there are sentences in which the object is replaced for the natural and atypical sentences.

\subsubsection{Extended calibration set}

We additionally generated a second set of calibration sentences for use in our experiment investigating distinctiveness (\Cref{sec:experiment-3-distinctiveness}).
In this experiment, we ask participants to judge pairs of sentences on the basis of the similarity of the events they describe.
A crucial feature that we aim for our manually generated pairs to have is that, in half the pairs, the verb contained in both sentences has the same sense, and in the other half, it has a clearly distinct sense.

We started with the natural, typical sentences written for our first set of calibration sentences. 
We then added to this set, for each natural and typical sentences, another sentence that used the \textit{same} sense as the natural and typical sentence, such as in \Cref{ex:second-manual-sentences-ex1}.

\ex. \label{ex:second-manual-sentences-ex1}
\a. The cook smashed the potatoes.
\b. The shoe smashed the bug.

We then generated another sentence that used what we deemed to be a different sense compared to that of the reference sentence, as in \Cref{ex:second-manual-sentences-ex2}.

\ex. \label{ex:second-manual-sentences-ex2}
\a. The cook smashed the potatoes.
\b. The startup smashed the competition.

This process results in 64 pairs of sentences, in which 32 are same-sense pairs and 32 are different-sense pairs.



\subsection{Sampling from a corpus}
\label{sec:implementing-sampling-from-a-corpus}

To implement sampling from a corpus, we sampled comments from Reddit found in the PushShift dataset \citep{baumgartner_pushshift_2020}.  
First, we automatically sense-tagged these raw sentences using a highly performant sense-tagger \citep{shi_simple_2019,orlando_transformer-srl_2020} trained on the PropBank annotations in the version of OntoNotes \citep{hovy_ontonotes_2006} used in the CoNLL-2012 shared task \citep{pradhan_conll-2012_2012}. 
For example, \Cref{ex:corpus-edit1-raw} from \Cref{sec:three-approaches-to-generating-examples} was sense-tagged to represent the \textit{turn to, go to} sense of \textit{hit} in PropBank, while \Cref{ex:corpus-edit2-raw} was sense-tagged as \textit{reach, encounter}.

\ex. \label{ex:corpus-edit2}
\a. \textbf{Raw:} Like the sensation was so strong, I went into what felt like shock, a tingle up my spine, and my body hit the floor.\label{ex:corpus-edit2-raw}
\b. \textbf{Post-edited:} The body hit the floor.\label{ex:corpus-edit2-postedit}

In total, we tagged approximately $\sim$129 million comments ($\sim$3.5 billion words).


As noted in \Cref{sec:three-approaches-to-generating-examples}, it is very rare to find sentences that perfectly satisfy the constraints in \Cref{ex:constraints}: 
many sentences are multi-clausal, use complex grammatical tense and/or aspect, and contain many pronouns. 
To deal with this issue, we use the SpaCy \citep[v3.5.3;][]{honnibal_spacy_2017} dependency parser (\texttt{en\_core\_web\_lg}) to filter data to sentences containing our target verbs whose dependents include both a subject and an object noun with determiners (excluding highly bleached noun phrases such as \textit{a lot}). 
We then automatically edit these sentences---changing all verb tenses to the past, replacing all determiners with \textit{the} and removing extraneous clauses to fit the desired form. 
\Cref{ex:corpus-edit2-postedit} shows the result of this procedure for \Cref{ex:corpus-edit2-raw}. 

The result of this editing procedure is not guaranteed to be a particularly natural sentence. 
As a heuristic for finding the most natural sentences, we computed each edited sentence's surprisal---i.e. the negative of its log-likelihood---using GPT-2 \citep{radford_language_2018} and manually reviewed the 10 lowest surprisal sentences exemplifying each sense, for all target verbs (1,032 sentences total). 
Of these, we disqualified sentences which were misparsed---e.g. \textit{B.C.} (British Columbia) in \Cref{ex:misparse}---incorrectly sense-tagged---e.g. \texttt{assert.02} assigned to \textit{belt} in \Cref{ex:mistagged}---or which used jargon/esoteric NPs---e.g. the proper name \textit{Kulaks} in \Cref{ex:esoteric}.

\ex.
\a. The B.C. burned the land.\label{ex:misparse}
\b. The fanbase belted the anthem.\label{ex:mistagged}
\b. The Kulaks burned the crops.\label{ex:esoteric}

The fact that we use a manual selection procedure means that the sentences are not selected purely automatically. 
Our aim in performing manual selection was to simulate the result of an automatic procedure whose inputs were perfectly annotated. 
Given that no such annotation is perfect, our results in \Crefrange{sec:experiment-1-naturalness}{sec:experiment-3-distinctiveness} must be interpreted as an upper bound on the quality of examples generated from a corpus. 

After this filtering step, we selected the four sentences for each sense with the lowest surprisal (based on the post-edited version of the sentences). 
If there were less than 4 sentences fitting these criteria, we took the top 2--3.\footnote{
    Some senses of verbs are only found in particular dialects and/or genres of English. 
    Reddit features primarily US English, and as a consequence, some rare senses of verbs are not represented at all within our data. For example, the verb \textit{pinch} has a common sense \texttt{pinch.01:\ squeeze tightly to cause pain} which is widely used by all speakers of English. A more uncommon sense is \texttt{pinch.02:\ slangy steal}, which is not common in US English. In the case where the LM-generated sentences represent such senses (described in \Cref{sec:generating-examples-from-a-language-model}) that Reddit did not, we simply do not use a Reddit parallel for those senses in our experiments.
} 
\Cref{tab:hitreddit} shows some examples of sentences representing three senses of \textit{hit}. 
Our final set of corpus-generated sentences comprises 367 sentences.

\begin{table}[t]
\centering
\caption{Examples for \textit{hit} found in sense-tagged data from Reddit.}
\label{tab:hitreddit}
\begin{tabular}{ll}
\toprule
     \textbf{PropBank sense gloss} & \textbf{Example from Reddit} \\
\midrule
    \texttt{hit.01} (strike) &   The bullets hit the wall. \\
   \texttt{hit.02} (reach, encounter)     &     The tape hit the news. \\
    \texttt{hit.03} (go to, turn to) & The resolutioners hit the gym. \\
    \bottomrule

\end{tabular}
\end{table}

\subsection{Sampling from a language model}
\label{sec:generating-examples-from-a-language-model}

To implement sampling from an LM, we sampled sentences from a quantized variant of \texttt{llama-2-13b}\footnote{
    A quantized variant of an LM is one that represents the parameters of the original LM using a lower precision numeric representation than the LM was originally trained with. 
    Quantization makes LMs substantially more efficient to sample from. Specifically, we used a model (\texttt{Q5\_K\_M} from \url{https://huggingface.co/TheBloke/Llama-2-13B-GGUF}) with 5-bit quantization (type 1) and super-blocks containing 8 blocks, each block having 32 weights and scales and mins quantized with 6 bits.
} using nucleus sampling \citep{holtzman_curious_2019}.\footnote{
   We used constrained sampling to produce at most 32 tokens and used a top-$p$ of 0.95, a min-$p$ of 0.05, a typical-$p$ of 1.0, a repeat penalty of 1.1, a tail-free sampling parameter of 1.0, a target cross-entropy of 5.0, a learning rate used to update mu of 0.1, a top-$k$ of 40, and starting temperature of 0.8 for each verb-sense. Most of these settings are the default from \url{https://llama-cpp-python.readthedocs.io/en/latest/api-reference/}.
    To generate unique sentences, we incremented the seed by 1 on successive calls to the sampler. 
    If after 100 increments in seed, a sufficient number of unique sentences had not been sampled---as noted below, we aimed for 10 for every verb-sense pair---the temperature was increased by 0.1 and sampling continued.
    This procedure was iterated until either a sufficient number of unique sentences were sampled or 100 different seeds at 100 different temperatures were attempted.
}
Sampling was also conditioned on the prompt template in \Cref{ex:prompt-template}, where \texttt{\{\{VERB\}\}} is replaced by the root form of the verb of interest and \texttt{\{\{SENSE\_GLOSS\}\}} is replaced by a gloss of one of that verb's senses, and was constrained by the grammar in \Cref{ex:grammar}, replacing the rule \texttt{V -> hit} with the simple past tense form of the verb of interest (see \Cref{sec:sampling-from-a-language-model}).  

\ex. An example of a sentence containing the verb ``\texttt{\{\{VERB\}\}}'' in the sense\\ ``\texttt{\{\{SENSE\_GLOSS\}\}}'':\label{ex:prompt-template} 

As in the manual and corpus-based generation methods, we aim to generate four sentences per verb sense. To achieve this, we sample 10 sentences per verb-sense then rank those sentences by surprisal.\footnote{
    We were unable to sample 10 unique sentences for the one sense of \textit{see}---\texttt{see.09} (\textit{visit/consultation by medical professional}) from Propbank---for which we could only obtain four unique sentences despite attempts using 100 different seeds at 100+ different temperatures.
} 
Then, we select the four with the lowest surprisal.
This approach was necessary in order to eliminate nonsense generations, such as \Cref{ex:nonsense}.\footnote{Nonsensical run-on sentences such as this arise because of the LM's strong proclivity towards a token that does not match our grammar, which the LM then forces to fit into our grammar by appending more tokens until it does. For this specific example, the fact that the probability of `.' is quite low after `salted' (related to it not being a noun) causes the LM to append tokens until the entire word is a noun.}

\ex. The waiter passed the saltedbuttertohiscustomerintheplasticdishandthencollecteditfromhimafterhewasdonewithit.\label{ex:nonsense}

We derive the sense glosses in one of two ways: (i) we use the sense glosses provided in PropBank directly; and (ii) we produce sense glosses from an LM.
The idea behind comparing these two alternatives is to understand how much one can rely solely on an LM to generate examples without the need for external resources like PropBank.
This setting is of particular interest for languages that do not have such sense lexicons---the vast majority of the world's languages.\footnote{
    This comparison is not perfect, since it is likely that sense lexicons like PropBank (and other dictionary resources) occur in many English LM's training data.
    Thus, insofar as such resources do not exist in some language of interest, an LM for that language will not have been trained on such resources, and sense generation methods based on them may behave differently than ones based on an English LM.
}

\begin{table}[t]
\centering
\caption{Examples for \textit{hit} given by the LM when prompted with PropBank's sense glosses.}
\label{tab:hitpb}
\begin{tabular}{ll}
\toprule
     \textbf{PropBank sense} & \textbf{Example from LM} \\
\midrule
     \texttt{hit.01} (strike) & The baseball hit the fence. \\
     \texttt{hit.02} (reach, encounter) & The ball hit the wall. \\
      \texttt{hit.03} (go to, turn to) & The road hit the lakefront. \\
\bottomrule
\end{tabular}
\end{table}

\Cref{tab:hitpb} shows example generations for the verb \textit{hit}, conditioned on the senses of \textit{hit} found in PropBank. To implement the second method of deriving sense glosses, we sampled sense glosses from a quantized version of \texttt{llama-2-13b-chat}.\footnote{We used the \texttt{gptq-8bit-128g-actorder\_True} from \url{https://huggingface.co/TheBloke/Llama-2-13B-chat-GPTQ}. We set the maximum tokens was set to 512, the minimum length to 10, the top-$k$ to 50, and try temperatures of 0.7, 0.8, and 0.9. Most of these parameters are the defaults from \url{https://llama-cpp-python.readthedocs.io/en/latest/api-reference/}.} 
This method consisted of two stages: (i) prompting the LM to classify verbs into monosemous and polysemous verbs; (ii) prompting the LM (a) to produce a single sense gloss for those verbs it classified as monsemous; or (b) to produce an enumeration of sense glosses for those verbs it classified as polysemous.  

We took this two-stage approach because we found that the LM tended to offer multiple senses for all verbs---even those listed as monosemous in PropBank.\footnote{
    One might posit that this pattern is evidence that PropBank ``undercounts'' senses---and that may well be the case \citep[see][]{petersen_lexical_2023}. But even if so, it seems likely that the number of senses posited by the LM for a verb would be at least moderately correlated with the number found in PropBank, and we found that it was not across the vast majority of draws from the LM.
}
This behavior is potentially problematic for comparing manually generated sense lexicons, like PropBank's, and sense lexicons generated from an LM because the quality of the LM-generated lexicon is likely highly sensitive to the form of the prompt and other factors---e.g. the prevalence and format of different kinds of sense lexicons and dictionaries in the data the LM was trained on.
We cannot completely mitigate these factors, but we attempt to control for at least some of them---specifically, the relative number of senses associated with a verb---by synthesizing the results across multiple prompts in a way that aligns best with PropBank.\footnote{
    This procedure is relatively generic and could be used to compare with alternative sense lexicons.
}
Our aim here is to focus our analysis mainly on the quality of the glosses produced by the LM, rather than the number.

\subsubsection{Stage 1: Monosemy v. polysemy}

To implement the first stage and hopefully rein in the LLM's tendency for sense proliferation, we derived a measure of monosemy/polysemy for each verb from the eight prompt template variants found in \Cref{ex:polysemy-prompt}. We then combined these measures to best predict monosemy/polysemy in PropBank.

\ex. \label{ex:polysemy-prompt}
\a. Does the verb ``\texttt{\{\{VERB\}\}}'' have only one sense when used in a transitive clause? 
\b. Does the verb ``\texttt{\{\{VERB\}\}}'' have only one possible meaning when used in a transitive clause? 
\b. Does the verb ``\texttt{\{\{VERB\}\}}'' have only one sense when used in a transitive clause? Only answer with YES or NO.
\b. Does the verb ``\texttt{\{\{VERB\}\}}'' have only one possible meaning when used in a transitive clause? Only answer with YES or NO.
\b. Does the verb ``\texttt{\{\{VERB\}\}}'' have MORE THAN one distinct meaning when used in a transitive clause?
\b. Does the verb ``\texttt{\{\{VERB\}\}}'' have more than one distinct meaning when used in a transitive clause?
\b. When used in a transitive clause, does the verb ``\texttt{\{\{VERB\}\}}'' have ONE meaning, or MORE THAN ONE distinct meaning?
\b. When used in a transitive clause, does the verb ``\texttt{\{\{VERB\}\}}'' have one meaning, or more than one distinct meaning?

We derived this measure by computing the log-odds $\log p(\text{\sc yes})-\log p(\text{\sc no})$, where $p(\text{\sc yes})$ is the probability that the first output token is \textit{yes} (or some capitalized variant)---and similarly for \textit{no}.
We then trained a support vector machine (SVM) to predict whether a verb is monosemous in PropBank using these eight measures as predictors, which we validated using nested cross-validation.\footnote{
    This approach splits the data into outer test folds and train-dev folds. 
    Iterating through each pairing of train-dev and test folds, we further split train-dev folds into inner development folds and train folds. 
    We train various model on train folds and evaluate on development folds to receive a score---performing grid search over kernel type (\texttt{linear}, \texttt{rbf} with \texttt{sklearn}'s default kernel coefficient) and regularization parameter ($0.01$, $0.1$, $1.0$, $2.0$, $5.0$, $10.0$)---and to select the best model. 
    The final score indicates performance on the outer folds of the model selected on the inner folds.
} 

This classifier achieves an accuracy of 0.55 in predicting whether a verb that was held-out in the cross-validation is monosemous or polysemous according to Propbank.\footnote{
    We provide this accuracy mainly as a point of information.
    Our goal in building this classifier is not high accuracy, but rather optimal alignment between the number of senses posited in a manually constructed lexicon and the number posited by the LM-based method.
    That optimal alignment may not be particularly good, as in this case.
}
We refit the classifier with the optimized hyperparameters to the entire dataset, then use this classifier to predict monosemy v. polysemy for use in the second stage.
The accuracy of this refit classifier is 0.77.\footnote{
    This accuracy is not informative of the classifiers performance, since it is evaluated on the data it was trained on. 
    Again, our goal here is optimal alignment with PropBank, not evaluation of the model on monosemy v. polysemy prediction.
}
In total, 30 verbs were labeled as monosemous (including \textit{allow}, \textit{box}, and \textit{button}) and 34 verbs were labelled as polysemous (including \textit{abuse}, \textit{beat}, and \textit{break}).

\subsubsection{Stage 2: Sense gloss generation}

We use the output of the classifier developed in the first stage to determine whether to generate one sense gloss for a verb or multiple. 
In cases, where we aim to generate just one gloss, we use the prompt in \Cref{ex:monosemous-prompt}.

\ex. Please describe the one possible sense of the verb ``\texttt{\{\{VERB\}\}}'' when it is used in a transitive clause.\label{ex:monosemous-prompt}

\begin{table}[t]
\centering
\caption{Examples for \textit{hit} given by the LLM with the LLM's own sense glosses.}
\label{tab:hitllm}
\begin{tabular}{ll}
\toprule
     \textbf{LM's own sense gloss} & \textbf{Example from LLM} \\
\midrule
     strike or collide with something & The ball hit the wall. \\  
     reach or affect something & The storm hit the coastline. \\ 
    play a musical note & The piano hit the notes. \\
     be popular or successful &  The song hit the charts. \\ 
      use violence or force & The police hit the rioters. \\ 
      cause a reaction or response & The movie hit the critics. \\ 
      fulfill a requirement or expectation & The candidate hit the mark. \\ 
\bottomrule
\end{tabular}
\end{table}

In cases where we need to generate multiple glosses, we again attempt to align the number of glosses with the number of glosses in PropBank. we start with the base prompt in \Cref{ex:base-prompt}. We then append to this prompt each possible combination of 1--4 prompts---always in the order given here. 

\ex. Describe and enumerate the distinct possible senses of the verb ``\texttt{\{\{VERB\}\}}'' when it is found in a transitive clause.\label{ex:base-prompt}

\ex. \label{ex:prompt-components}
\a. For example, if you were given the verb ``administrate'', you should respond with ``manage'' because ``administrate'' has one transitive sense.
\b. For example, if you were given the verb ``abandon'', you should respond with ``1. leave behind; 2. exchange; 3. surrender, give over'' since ``abandon'' has three transitive senses.
\b. For example, the verb ``jump'' has five senses because it has multiple possible meanings when it is used, so you should output something like ``1. stock prices, increase, 2. be excited for an opportunity, getting there first, 3. physically or metaphorically leap, physical motion, 4. to escape, bail out, 5. attack, gangsta style''.
\b. Ensure that the sense description(s) can stand alone and do not depend on being the synonym of some other verb.

The examples of sense descriptions are taken verbatim from a verb in Propbank for which we are not generating senses in order to better align the sense lexicons. 

We parse the output of these prompts to retrieve an integer value for each prompt-verb pair. 
For each prompt, we calculate the mean-absolute error between its sense counts compared to Propbank's to select the prompt whose sense count distribution best matches Propbank's. 
We use bootstrapping to estimate confidence intervals for the mean absolute error differences between the best prompt and other prompts. 
This gives us a set of best prompts, of which we choose the one with the shortest length and lowest temperature. 
This yields the final prompt in \Cref{ex:final-polysemy-prompt}.

\ex. Describe and enumerate the distinct possible senses of the verb ``\texttt{\{\{VERB\}\}}'' when it is found in a transitive clause. 
Feel free to give only one sense if it only has one possible meaning. 
For example, if you were given the verb ``administrate'', you should respond with ``manage'' because ``administrate'' has one transitive sense.\label{ex:final-polysemy-prompt}     

We suspect that \Cref{ex:final-polysemy-prompt} is successful because its only example is a verb said to be monosemous, combating the LM's tendency to offer many sense glosses.
Using this prompt, our LM gives a median of one sense to the verbs that PropBank categorizes as monosemous, and a median of five senses to those that PropBank categorizes as polysemous (compared to PropBank's median of three senses for such verbs). \Cref{tab:hitllm} shows example generations for the verb \textit{hit}, conditioned on the LM-generated senses of \textit{hit}, while \Cref{tab:hitllmpropbank} shows example generations for the same verb conditioned on Propbank's sense glosses.


\begin{table}[t]
\centering
\caption{Examples for \textit{hit} given by the LLM with Propbank's transitive sense glosses.}
\label{tab:hitllmpropbank}
\begin{tabular}{ll}
\toprule
     \textbf{Propbank's sense gloss} & \textbf{Example from LLM} \\
\midrule 
     hit.01: strike & The pitcher hit the ball. \\  
    hit.02: reach, encounter & 
     The ball hit the wall. \\ 
    hit.03: go to, turn to & The road hit the town. \\
\bottomrule
\end{tabular}
\end{table}

\section{Experiment 1: Naturalness}
\label{sec:experiment-1-naturalness}

Our naturalness experiment was separated into two subexperiments: one focused on the manually generated sentences (\Cref{sec:subexperiment-1-1-automatically-generated-sentences}) and the other focused on the automatically generated sentences (\Cref{sec:subexperiment-1-2-automatically-generated-sentences}).

\subsection{Instructions and practice sentences}
\label{sec:exp1-instructions-and-practice-sentences}

In both subexperiments, participants are asked to rate the naturalness of each sentence on a continuous scale ranging from \textit{extremely unnatural} to \textit{perfectly natural}, where naturalness is defined for participants in the following way: ``a natural sentence is something that a native speaker of English would naturally and fluently say, following the implicit structural rules of English.''
As an example, participants are told that \Cref{ex:sleep1} is natural, \Cref{ex:sleep2} is ``somewhat natural, because---even though it doesn't make sense (toothbrushes don't sleep)---it does follow the structural rules of English'' and \Cref{ex:sleep3} is unnatural. 

\ex. \label{ex:sleep} 
\a. \label{ex:sleep1} The baby seems to be sleeping.
\b. \label{ex:sleep2} The toothbrush seems to be sleeping. 
\c.  \label{ex:sleep3} The baby seems sleep to be.

Participants are then given three practice sentences similar to those in \Cref{ex:sleep}. 
Participants are given feedback on these practice sentences, which they have to rate correctly---rating the sentences like \Cref{ex:sleep1}--\Cref{ex:sleep2} above the midpoint, and the one like \Cref{ex:sleep3} below the midpoint---in order to continue. Participants that did not correctly complete the practice sentences within two attempts were not allowed to continue. 

\begin{figure}
    \centering
    \includegraphics[width=\textwidth]{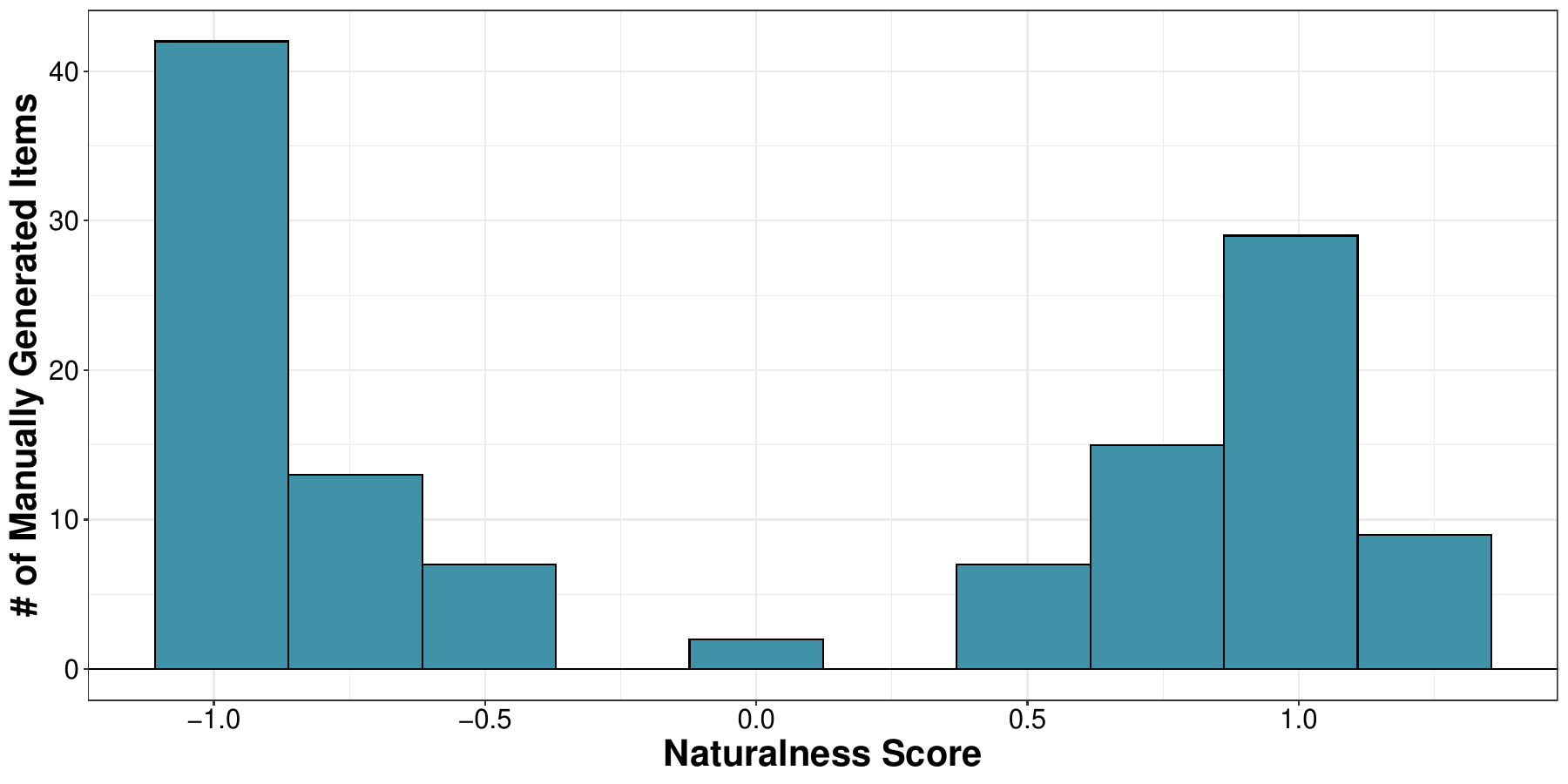}
    \caption{Distribution of naturalness scores for manually generated sentences.}
    \label{fig:hist-naturalness-zscore}
\end{figure}

\subsection{Subexperiment 1.1: Manually generated sentences}
\label{sec:subexperiment-1-1-automatically-generated-sentences}

Subexperiment 1.1 has two purposes: 
(i) to assess the extent to which experts can reliably generate natural and unnatural examples; 
and (ii) to produce a set of non-target examples beyond those discussed in \Cref{sec:exp1-instructions-and-practice-sentences} to be used in calibrating participants to the rating scale in Subexperiment 1.2.

\subsubsection{Materials}
The process for generating 124 manually-selected sentences is detailed in Section \Cref{sec:manually-generating-examples}. Participants rated all sentences in a randomized order after a practice block of 3 questions.

\subsubsection{Participants}

We recruited 55 participants to rate our manually-generated sentences, all who passed our practice questions, with the goal of having each sentence rated by at least 5 annotators. All annotators were self-identified native English speakers located in the United States and recruited through the Prolific web platform.  No annotator was allowed to participate in more than one of these studies.  With IRB approval from our institutions, annotators were paid an average of \$12/hour.

\subsubsection{Selecting calibration sentences}\label{sec:selecting-calibration-sentences}

We selected 50 of the 124 manually generated sentences to use as calibration sentences in Subexperiment 1.2. The point of these calibration sentences is to ensure that the responses participants provide in Subexperiment 1.2 are comparable to those that we observe in Subexperiment 1.1. Ensuring this comparability is crucial, since if we only provided participants with the automatically generated sentences, they may calibrate to the variability in naturalness found among those sentences, making the ratings incomparable to the ratings for the manually generated sentences.

In selecting calibration sentences, we aim for a subset of sentences (i) in which the ratings are as uniformly distributed as possible; (ii) in which the mean rating is as close as possible to the mean across all sentences; (iii) that contains sentences headed by the same verb no more than twice; and (iv) for verbs found in two sentences, the difference in the naturalness ratings for that verb's sentences is at least 0.5 standard deviations.
To satisfy these criteria, we first $z$-scored the naturalness ratings by participant in order to normalize for differences in scale use and took the mean of these $z$-scored ratings by sentence to derive a single naturalness score for each sentence.
\Cref{fig:hist-naturalness-zscore} shows the distribution of naturalness scores.
The bimodal distribution observed in \Cref{fig:hist-naturalness-zscore} is expected, given that we engineered sentences to be either natural or unnatural.

We then selected sentences for inclusion by (i) sorting sentences by their absolute distance to the mean naturalness score across sentences; (ii) moving through the sort, keeping sentences that move the mean of the included sentences toward the mean across all sentences and rejecting sentences that either (a) move the subset mean in the wrong direction; or (b) that would violate the fourth constraint described above.

Of this subset of 50 sentences, the eight sentences that were selected first in the procedure described above were reserved as the first sentences that participants saw in Subexperiment 1.2 (before any target sentences were shown and after the guided questions and introduction). 
We refer to this set as the \textit{initial calibration set}, used in a \textit{calibration block}. 
The remainder were pseudorandomly interleaved with the target sentences in Subexperiment 1.2 in their specified order to ensure that participants remain calibrated over the course of the experiment.

\begin{table}[t]
\footnotesize
\centering
 \caption{Most and least natural sentences by generation procedure, according to our participants' $z$-scored ratings.}
 \label{tab:naturalnessexample}
\begin{tabular}{lll}
\toprule
 & \textbf{Most natural}  & \textbf{Least natural}  \\
 \midrule
\multirow{ 1}{*}{Corpus} & The people knew the truth.  
 & The leviathan encountered the nothing.\\
\midrule
\multirow{ 1}{*}{LM + PropBank senses} & The company fired the employee.  
& The letter posted the previous.  \\
\midrule
\multirow{ 1}{*}{LM + LM senses} & The police hit the protesters. 
& The movie drew the entiretyofmyattention. \\
\bottomrule
 \end{tabular}
\end{table}

\subsection{Subexperiment 1.2: Automatically generated sentences}
\label{sec:subexperiment-1-2-automatically-generated-sentences}

Although we technically had three different naturalness experiments, one per automatic generation method, we consider all the data together and treat the different methods as factors between subjects.

\subsubsection{Materials}\label{sec:subexperiment-1-2-automatically-generated-examples:materials}
The process for automatically generating corpus-based sentences with Propbank senses, LM sentences with Propbank senses, and LM sentences with LM senses is detailed in \Cref{sec:three-approaches-to-generating-examples}. Since there were over 300 sentences generated per process, we used lists to allocate a reasonable number of sentences per participant. In order to generate lists of target sentences interwoven with calibration sentences, we first assigned each target sentence to a random list number (with a seed of 0) in ascending order such that there are less than 60 target sentences with the same list number. After splitting these sentences into their allocated list, we add calibration sentences in the order in which we specified in \Cref{sec:selecting-calibration-sentences} for a total of 89 sentences. Since there are already 3 practice sentences, and each survey also had a block of 8 calibration sentences shown after the instructions and practice sentences but before the target sentences, this leads to a total of 100 sentences per survey/list. 
There were 7 lists generated for the survey with corpus-based sentences, 10 lists generated for the survey with LM sentences with Propbank senses, and 15 lists generated for the survey using LM sentences with LM senses.

\subsubsection{Participants}

We recruited 61 participants to rate corpus-generated sentences of which 60 participants passed the practice questions; 62 participants to rate sentences generated using the LM prompted with PropBank senses of which 61 participants passed the practice questions; and 86 participants to rate sentences generated using the LM prompted with LM senses of which 79 participants passed the practice questions. 

\begin{figure}[t]
  \centering
\includegraphics[width=\textwidth]{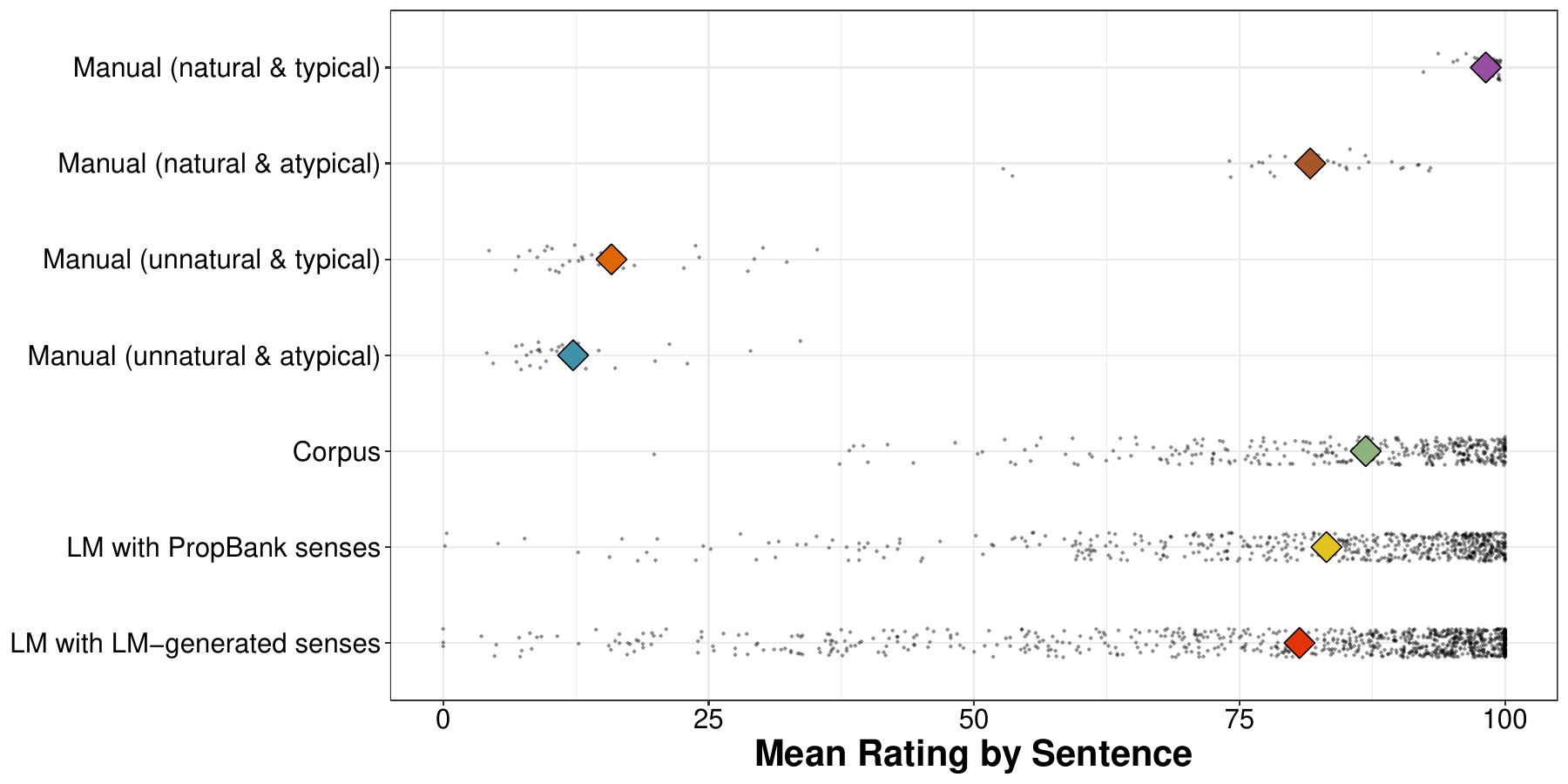}
  \caption{Mean naturalness rating for each sentence produced by each generation method. Each black point shows the mean rating of a sentence and large colored points show the mean of those means for each generation method.}
  \label{fig:mean-nat}
\end{figure}

Every sentence was rated by at least 5 annotators, all self-identified native English speakers located in the United States and recruited through the Prolific web platform.  No annotator was allowed to participate in more than one of these studies.  With IRB approval from our institutions, annotators were paid an average of \$12/hour.

\subsection{Results}

\Cref{fig:mean-nat} shows the mean naturalness ratings for each sentence from each generation method. Each black point shows the mean rating of a sentence and large colored points show the mean of those means. For examples of these sentences, \Cref{tab:naturalnessexample} provides the highest and lowest rated items by generation procedure.
As expected, the manually generated examples that were both natural and typical are indeed rated as very natural. 
The examples that natural but atypical receive slightly lower ratings, potentially suggesting that naturalness is influenced by typicality. 
But this effect is nowhere near as strong as the effect of being manually engineered to be unnatural: 
all of the examples that were engineered to be unnatural are rated as such. 

The automatically generated examples show naturalness ratings that tend to be about as good as---or slightly better than---the manually generated sentences constructed to be natural but atypical. 
We hypothesize that these ratings are a product of the automatically generated examples being less typical than the manually generated sentences that were engineered to be both natural and typical---a hypothesis we return to in \Cref{sec:experiment-2-typicality}

\subsection{Analysis}

To assess the reliability of the differences observed in \Cref{fig:mean-nat}, we fit an ordered beta mixed effects model \citep{kubinec_ordered_2023} to the responses.\footnote{
    This fit and all other ordered beta mixed model fits discussed in the paper are implemented with Hamiltonian Monte Carlo in STAN \citep{carpenter_stan_2017}, using the \texttt{cmdstanpy} interface to \texttt{cmdstan}.
    All fits include four chains of 5,000 samples each (2,500 of which are treated as warmup).
    All fits pass all default diagnostic checks implemented in STAN.
} 
This model had fixed effects for generation method---treating classes of manually generated sentences as separate generation methods---as well as by-participant, by-verb, by-sense, and by-sentence random intercepts. \Cref{tab:naturalness-model-base} shows the estimates for the fixed effect coefficients. All automatic generation methods produce reliably less natural examples than manual generation, though all such methods produce examples that are likely more natural than the manually generated natural, atypical examples: corpus (posterior $p > $ 0.99), LM with PropBank senses (posterior $p = $ 0.94), and LM with LM senses (posterior $p = $ 0.97).\footnote{
    In general, we refer to an effect as \textit{reliable} if the posterior probability that the effect has a different sign than its posterior mean is $\leq$1\%.
    This terminology is meant to be analogous to terminology common in frequentist null hypothesis testing---though we stress that, because we are working in a Bayesian paradigm, our use of the term is merely heuristic.
}

\begin{table}[t]
    \centering
    \small
    \caption{Fixed effect coefficient estimates in log-odds space for naturalness experiment with \textit{Manual (Natural \& Typical)} as the reference-level in a dummy coding. The 2.5\% and 97.5\% columns give the lower and upper bound of the 95\% credible interval, respectively, and the posterior $p$ column gives the posterior probability that the coefficient has a sign different from the posterior mean. The posterior mean of the lower cutpoint of the ordered beta model is $-$1.99 (95\% CI = [$-$2.04, $-$1.94]) and the posterior mean of the mean of the upper cutpoints is 0.42 (95\% CI = [0.15, 0.74]).}
    \label{tab:naturalness-model-base}
\begin{tabular}{rrrrr}
\toprule
      & \textbf{Post. mean} & \textbf{2.5\%} & \textbf{97.5\%} & \textbf{Post.} $p$ \\
\midrule
Intercept & 2.95 & 2.61 & 3.30 & $<$ 0.01 \\
Manual (Natural \& Atypical) & -2.15 & -2.50 & -1.81 & $<$ 0.01 \\
Manual (Unnatural \& Typical) & -4.52 & -4.87 & -4.18 & $<$ 0.01 \\
Manual (Unnatural \& Atypical) & -4.76 & -5.10 & -4.41 & $<$ 0.01 \\
Corpus & -1.71 & -2.08 & -1.33 & $<$ 0.01 \\
LM with manually-generated sense glosses & -1.88 & -2.25 & -1.52 & $<$ 0.01 \\
LM with LM-generated sense glosses & -1.84 & -2.20 & -1.48 & $<$ 0.01 \\
\bottomrule
\end{tabular}
\end{table}

\subsection{Discussion}

We find that examples produced by the automated methods we consider are not as natural as our best manually generated sentences but that they are nonetheless quite natural---being rated as more natural, on average, than sentences we manually generated to be natural but atypical.
As discussed in \Cref{sec:three-approaches-to-generating-examples}, this pattern could arise because both the corpus-sampling and LM-sampling methods are in principle sensitive to a verb's frequency: the less frequent the verb, the harder it will be to find examples of that verb that satisfy the constraints of inference in some corpus due to power laws. This sensitivity certainly affects the corpus-sampling method, but it could also affect the LM-sampling methods, given that they compress a corpus. How much it affects the latter is largely dependent on the nature of the abstractions is learns and how much their generalization capabilities are able to overcome power laws.

To assess whether frequency-sensitivity drives the results we observe, we obtain the frequency of each verb in our dataset when found in a transitive clause from the VALEX dataset \citep{korhonen-etal-2006-large}. 
We then fit ordered beta mixed effects model to the responses for the automatic generation methods---also including responses for the natural, typical subset of the manual generation method for comparison, since we do not expect manual generation to be frequency sensitive. 
This model has fixed effects for generation method (\textit{manual} v. \textit{corpus} v. \textit{LM} collapsing across sense-generation method), $z$-scored frequency, and their multiplicative interaction as well as by-participant, by-verb, by-sense, and by-sentence random intercepts and by-participant random slopes for $z$-scored frequency.\footnote{
    We use the $z$-scores of the raw frequency obtained from VALEX, rather than the log of that frequency (as would be more standard), because we found that models using raw frequency produced better fits. 
}
Insofar as our automatic generation methods are frequency-sensitive, we expect a negative simple effect for those methods---bringing the average rating for lower frequency items down---as well as a positive interaction between frequency and each automated method---bring the average rating for higher frequency items up.

\begin{table}[t]
    \centering
    \small
    \caption{Fixed effect coefficient estimates in log-odds space for naturalness experiment with \textit{Manual} as the reference-level in a dummy coding. The 2.5\% and 97.5\% columns give the lower and upper bound of the 95\% credible interval, respectively, and the posterior $p$ column gives the posterior probability that the coefficient has a sign different from the posterior mean. The posterior mean of the lower cutpoint of the ordered beta model is $-$2.43 (95\% CI = [$-$2.53, $-$2.32]) and the posterior mean of the mean of the upper cutpoints is 0.05 (95\% CI = [$-$0.25, 0.37]).}
    \label{tab:naturalness-model-freq}
\begin{tabular}{rrrrr}
\toprule
      & \textbf{Post. mean} & \textbf{2.5\%} & \textbf{97.5\%} & \textbf{Post.} $p$ \\
\midrule
Intercept & 2.26 & 1.89 & 2.64 & $<$ 0.01 \\
Corpus & -0.89 & -1.33 & -0.47 & $<$ 0.01 \\
LM & -0.97 & -1.35 & -0.58 & $<$ 0.01 \\
Frequency ($z$-scored) & 0.20 & -0.14 & 0.54 & 0.13 \\
Corpus $\times$ Frequency ($z$-scored) & -0.09 & -0.46 & 0.28 & 0.32 \\
LM $\times$ Frequency ($z$-scored) & -0.10 & -0.45 & 0.26 & 0.30 \\
\bottomrule
\end{tabular}
\end{table}

\Cref{tab:naturalness-model-freq} shows the estimates for the fixed effect coefficients of this model. 
Consistent with our earlier analysis showing that the automated methods produce reliably less natural sentences, we find that the simple effect of both automated method types is reliably negative.
But inconsistent with the automated methods being frequency sensitive, we observe negative interactions between the automated methods and frequency.
These interactions are furthermore quite small relative to the simple effects of each automated method and not reliable.
Thus, when considering the naturalness of the expressions they produce, neither automated method appears to be more frequency-sensitive compared to manual generation.\footnote{
    We do observe a positive simple effect of frequency, potentially suggesting that naturalness ratings for our manually generated items are somewhat frequency-sensitive, but this effect is weak relative to the size of the intercept term and not reliable.
}

What then drives the lower naturalness ratings for the automated methods?
We hypothesize that these ratings are a product of the automatically generated examples being less typical than the manually generated sentences that were engineered to be both natural and typical. 
We investigate this hypothesis in \Cref{sec:experiment-2-typicality} but preliminarily conclude from the pattern observed in Experiment 1 that all of the automated generation methods we consider are safe to use for generating examples for downstream annotation insofar as some small amount of degradation in naturalness is tolerable.

\section{Experiment 2: Typicality}
\label{sec:experiment-2-typicality}

Like the naturalness experiment, our typicality experiment was separated into two subexperiments: one focused on the manually generated sentences (\Cref{sec:subexperiment-2-1-automatically-generated-sentences}) and the other focused on the automatically generated sentences (\Cref{sec:subexperiment-2-2-automatically-generated-sentences}). 

\subsection{Instructions and practice sentences}

In both subexperiments, participants are asked to rate the typicality of the event described by a sentence on a continuous scale ranging from \textit{very atypical} to \textit{very typical}.
They are instructed that ``a typical situation is one in which individuals and their actions follow our normal expectations.''  
They are told that \Cref{ex:typicality-example-typical} describes a typical situation, while \Cref{ex:typicality-example-less-typical} is less typical because ``waitresses typically serve food instead of cooking it and salads are not cooked;'' and \Cref{ex:typicality-example-least-typical} is the least typical of all because ``aliens are not typically associated with cooking and pencils are not things that are typically cooked.'' 

\ex. \label{ex:typicality-example}
\a. The chef cooked the meal.\label{ex:typicality-example-typical}
\b. The waitress cooked the salad.\label{ex:typicality-example-less-typical} 
\c. The alien cooked the pencil.\label{ex:typicality-example-least-typical} 

Participants are then given three practice sentences similar to \Cref{ex:typicality-example}. 
Participants are given feedback on these practice sentences, which they have to rate correctly---rating the sentences like \Cref{ex:typicality-example-typical}--\Cref{ex:typicality-example-less-typical} above the midpoint, and the one like \Cref{ex:typicality-example-least-typical} below the midpoint---in order to continue. 

\subsection{Subexperiment 2.1: Manually generated sentences}
\label{sec:subexperiment-2-1-automatically-generated-sentences}

\subsubsection{Materials}
The process for generating manually-selected sentences is detailed in Section \Cref{sec:manual-generation-by-experts}. Similar to Subexperiment 1.1, all sentences were shown to participants in a randomized order, after a practice block of 3 questions.

\subsubsection{Participants}

Similarly to Subexperiment 1.1, we recruited 55 participants from the same pool (although no participants participated in more than one of our subexperiments). They were compensated at the same rate.

\subsubsection{Selecting calibration senteces}
Our procedure for selecting calibration sentences for our future typicality subexperiments is directly analagous to the procedure described in \Cref{sec:selecting-calibration-sentences}.

\subsection{Subexperiment 2.2: Automatically generated sentences}
\label{sec:subexperiment-2-2-automatically-generated-sentences}

\subsubsection{Materials}
The procedure for generating the automatically generated sentences are detailed in \Cref{sec:implementing-the-three-approaches}. Similar to Subexperiment 1.2, since these processes yielded too many sentences to all be shown to participants in a single survey, we used the same lists generated in \Cref{sec:subexperiment-1-2-automatically-generated-examples:materials}.

\subsubsection{Participants}

We recruited 60 participants to rate corpus-generated sentences of which all 60 participants passed the practice questions; 62 participants to rate sentences generated using the LM prompted with PropBank senses of which 60 participants passed the practice questions; and 79 participants to rate sentences generated using the LM prompted with LM senses of which all 79 participants passed the practice questions. 

Every sentence was rated by at least 5 annotators, all self-identified native English speakers located in the United States and recruited through the Prolific web platform.  No annotator was allowed to participate in more than one of these studies.  With IRB approval from our institutions, annotators were paid an average of \$12/hour.

\subsection{Results}

\begin{table}[t]
\footnotesize
\centering
 \caption{Most and least typical sentences by generation procedure, according to our participants' $z$-scored ratings.}
 \label{tab:typicalityexample}
\begin{tabular}{lll}
\toprule
 & \textbf{Most typical}  & \textbf{Least typical}  \\
 \midrule
\multirow{ 1}{*}{Corpus} & The staff cleaned the tables. 
& The sign recalled the story.\\
\midrule
\multirow{ 2}{*}{LM + PropBank senses} & \multirow{ 2}{*}{The criminal broke the law.} & The chicken chopped the onions. \\ 
& & The water dropped the bread. \\
\midrule
\multirow{ 2}{*}{LM + LM senses} & The firemen saved the building. & \multirow{ 2}{*}{The verb phoned the verb.} \\ 
& The woman understood the idea.  &   \\
\bottomrule
 \end{tabular}
\end{table}

\begin{figure}[t]
  \centering
\includegraphics[width=\textwidth]{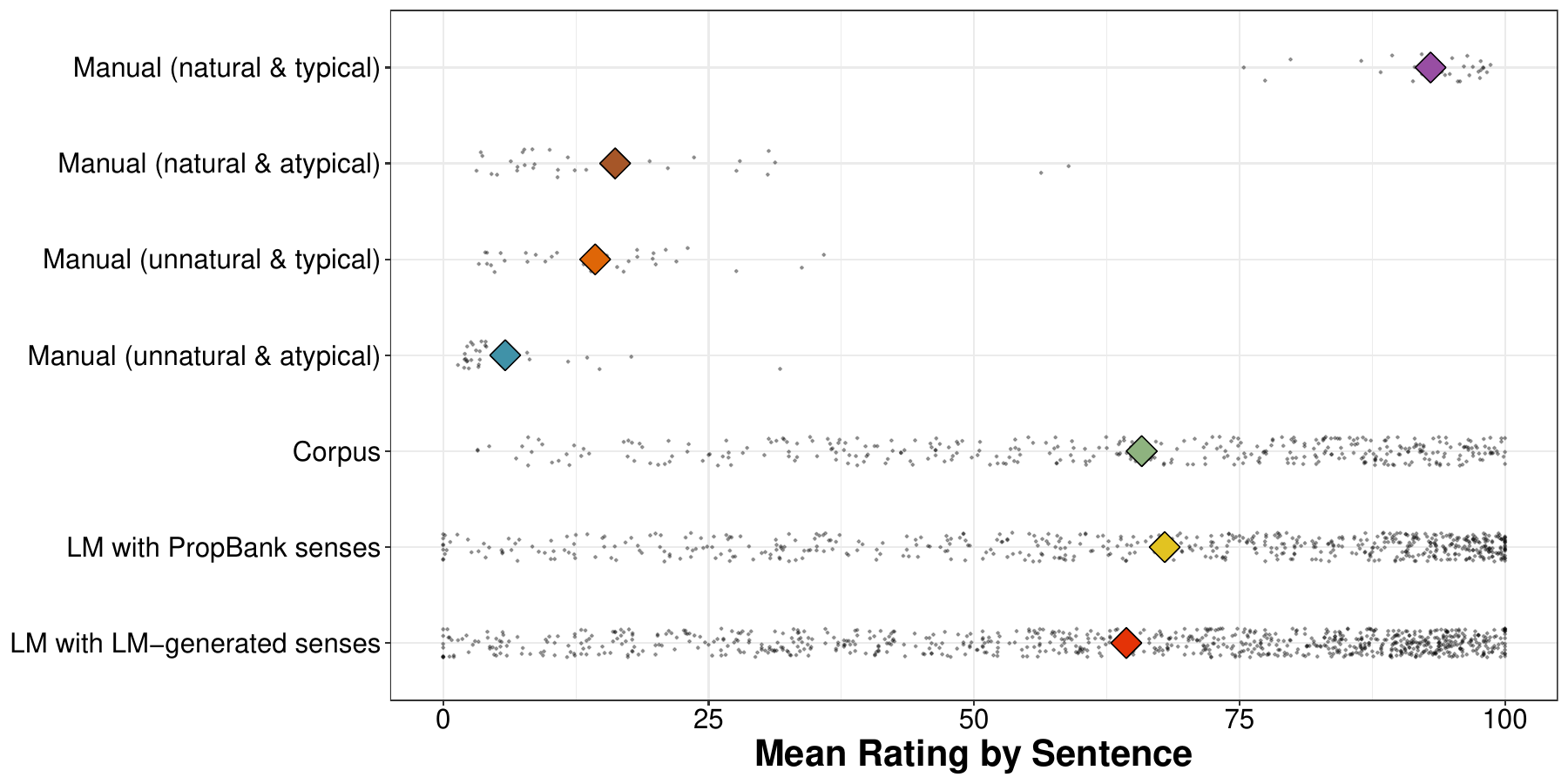}
  \caption{Mean typicality rating for each sentence from each source. Each black point shows the mean rating of a sentence and large colored points show the mean of those means for each generation method.}
  \label{fig:mean-typ}
\end{figure}

\Cref{fig:mean-typ} shows the mean typicality ratings for each sentence from each generation method. Each black point shows the mean rating of a sentence and large colored points show the mean of those means. For examples of these sentences, \Cref{tab:typicalityexample} provides the highest and lowest rated items by generation procedure.
As expected, the manually generated sentences that were constructed to be both natural and typical are indeed rated as very typical. 
All other categories of manually generated sentences received very low typicality ratings on average---including the sentences that were constructed to be unnatural but typical.  
Thus, it appears that naturalness exerts a strong effect on typicality---with only natural sentences being candidates for typical event descriptions. 

The automatically generated examples are rated as less typical than the manually generated sentences that are constructed to be natural and typical, though they are all rated to be substantially more typical than any of the other categories of manually generated sentences. 

\subsection{Analysis}

To assess the reliability of the differences observed in \Cref{fig:mean-typ}, we fit an ordered beta mixed effects model to the responses. 
This model had fixed effects for generation method---treating classes of manually generated sentences as separate generation methods---as well as by-participant, by-verb, by-sense, and by-sentence random intercepts. \Cref{tab:typicality-model-base} shows the estimates for the fixed effect coefficients. Consistent with the pattern observed in \Cref{fig:mean-typ}, all automatic generation methods produce reliably less typical event descriptions than manual generation, though all such methods produce examples that are more typical than the manually generated sentences that are constructed to be unnatural or atypical (all posterior $p$s $<$ 0.01).

\begin{table}[t]
    \centering
    \small
    \caption{Fixed effect coefficient estimates in log-odds space for typicality experiment with \textit{Manual (Natural \& Typical)} as the reference-level in a dummy coding. The 2.5\% and 97.5\% columns give the lower and upper bound of the 95\% credible interval, respectively, and the posterior $p$ column gives the posterior probability that the coefficient has a sign different from the posterior mean. The posterior mean of the lower cutpoint of the ordered beta model is $-$1.86 (95\% CrI = [$-$1.91, $-$1.82]) and the posterior mean of the mean of the upper cutpoints is 1.11 (95\% CrI = [0.82, 1.45]).}
    \label{tab:typicality-model-base}
\begin{tabular}{rrrrr}
\toprule
      & \textbf{Post. mean} & \textbf{2.5\%} & \textbf{97.5\%} & \textbf{Post.} $p$ \\
\midrule
Intercept & 1.82 & 1.43 & 2.21 & $<$ 0.01 \\
Manual (Natural \& Atypical) & -3.44 & -3.86 & -3.02 & $<$ 0.01 \\
Manual (Unnatural \& Typical) & -3.60 & -4.02 & -3.17 & $<$ 0.01 \\
Manual (Unnatural \& Atypical) & -4.31 & -4.73 & -3.89 & $<$ 0.01 \\
Corpus & -1.39 & -1.81 & -0.95 & $<$ 0.01 \\
LM with manually-generated sense glosses & -1.35 & -1.76 & -0.92 & $<$ 0.01 \\
LM with LM-generated sense glosses & -1.48 & -1.89 & -1.06 & $<$ 0.01 \\
\bottomrule
\end{tabular}
\end{table}

\subsection{Discussion}

We find that event descriptions conveyed by sentences produced using automatic methods we consider are not as typical as our best manually generated sentences but that they are nonetheless substantially more typical than event descriptions manually constructed to be atypical. 
We conclude from this result that all of the automated generation methods we consider are safe to use for generating sentences for downstream annotation insofar as some small amount of degradation in typicality is tolerable.

As discussed in \Cref{sec:three-approaches-to-generating-examples}, this pattern could arise because both the corpus-sampling and LM-sampling methods are in principle sensitive to a verb's frequency.
In \Cref{sec:experiment-1-naturalness}, we found that automated generated methods were no more frequency-sensitive than manual generation methods in terms of naturalness;
but in principle, they could be more frequency-sensitive in terms of typicality.

To assess whether frequency-sensitivity drives the typicality results we observe, we fit ordered beta mixed effects model to the typicality responses for the automatic generation methods---also including responses for the natural, typical subset of the manual generation method for comparison, since we do not expect manual generation to be frequency sensitive. 
Like the analogous model for naturalness, this model has fixed effects for generation method (\textit{manual} v. \textit{corpus} v. \textit{LM} collapsing across sense-generation method), $z$-scored frequency, and their multiplicative interaction as well as by-participant, by-verb, by-sense, and by-sentence random intercepts and by-participant random slopes for $z$-scored frequency.
And as for naturalness, insofar as our automatic generation methods are frequency-sensitive, we expect a negative simple effect for those methods---bringing the average rating for lower frequency items down---as well as a positive interaction between frequency and each automated method---bring the average rating for higher frequency items up.

\begin{table}[t]
    \centering
    \small
    \caption{Fixed effect coefficient estimates in log-odds space for typicality experiment with \textit{Manual} as the reference-level in a dummy coding. The 2.5\% and 97.5\% columns give the lower and upper bound of the 95\% credible interval, respectively, and the posterior $p$ column gives the posterior probability that the coefficient has a sign different from the posterior mean. The posterior mean of the lower cutpoint of the ordered beta model is $-$2.23 (95\% CI = [$-$2.3, $-$2.16]) and the posterior mean of the mean of the upper cutpoints is 0.79 (95\% CI = [0.50, 1.11]).}
    \label{tab:typicality-model-freq}
\begin{tabular}{rrrrr}
\toprule
      & \textbf{Post. mean} & \textbf{2.5\%} & \textbf{97.5\%} & \textbf{Post.} $p$ \\
\midrule
Intercept & 1.71 & 1.30 & 2.12 & $<$ 0.01 \\
Corpus & -1.34 & -1.79 & -0.89 & $<$ 0.01 \\
LM & -1.23 & -1.66 & -0.81 & $<$ 0.01 \\
Frequency ($z$-scored) & 0.19 & -0.26 & 0.63 & 0.20 \\
Corpus $\times$ Frequency ($z$-scored) & -0.11 & -0.59 & 0.36 & 0.32 \\
LM $\times$ Frequency ($z$-scored) & -0.02 & -0.48 & 0.44 & 0.46 \\
\bottomrule
\end{tabular}
\end{table}

\Cref{tab:typicality-model-freq} shows the estimates for the fixed effect coefficients of this model. 
Consistent with our earlier analysis showing that the automated methods produce reliably less typical sentences, we find that the simple effect of both automated method types is reliably negative.
But as for naturalness and inconsistent with the automated methods being frequency sensitive, we observe negative interactions between the automated methods and frequency.
These interactions are furthermore quite small relative to the simple effects of each automated method and not reliable.
Thus, as for naturalness, when considering the typicality of the expressions they produce, neither automated method appears to be more frequency-sensitive compared to manual generation.\footnote{
    As with naturalness, we do observe a positive simple effect of frequency, potentially suggesting that naturalness ratings for our manually generated items are somewhat frequency-sensitive; but similar to naturalness, this effect is weak relative to the size of the intercept term and not reliable.
}

In \Cref{sec:experiment-1-naturalness}, we hypothesized that typicality may play a role in the degraded naturalness of the automatically generated sentences.
To test this hypothesis, we derived a measure of typicality for each sentence by computing its average typicality rating $z$-scored by participant. 
We fit an ordered beta mixed effects model to the naturalness ratings reported on in \Cref{sec:experiment-1-naturalness} with this rating as a fixed effect as well as by-participant, by-verb, by-sense, and by-sentence random intercepts.
We find a strong, reliably positive effect of typicality ($\beta=$1.47, 95\% CrI=[1.33, 1.61])---consistent with typicality exerting a strong influence on naturalness ratings.
This result is consistent with the degradation in naturalness observed among the automatically generated sentences being driven mainly by typicality effects, rather than syntactic ill-formedness.

\section{Experiment 3: Distinctiveness}
\label{sec:experiment-3-distinctiveness}

Like the naturalness and typicality experiments, our distinctiveness experiment was separated into two subexperiments: one focused on the manually generated examples (\Cref{sec:subexperiment-3-1-manually-generated-sentences}) and the other focused on the automatically generated examples (\Cref{sec:subexperiment-3-2-automatically-generated-sentences}). 

\subsection{Instructions and practice sentences}

Participants were presented with two sentences describing situations and asked ``how similar or different the situations are compared to each other'' on a slider scale ranging from \textit{completely identical} to \textit{extremely different}.\footnote{
    This task is similar in design to \citeauthor{erk_investigations_2009}'s (\citeyear{erk_investigations_2009}) Usage Similarity task, which she uses to explore graded meaning similarity of target words. 
    The main difference is that we use a slider, rather than an ordinal scale.
}
As an example, they were given the pair in \Cref{ex:distinctiveness-instructions} and told \Cref{ex:distinctiveness-instructions-run1} and \Cref{ex:distinctiveness-instructions-run2} are extremely different because the former involves the physical movement of an athlete participating in a race, while the latter involves the managing and leadership of a company.

\ex. \label{ex:distinctiveness-instructions}
\a. The athlete ran the race.\label{ex:distinctiveness-instructions-run1}
\b. The manager ran the organization.\label{ex:distinctiveness-instructions-run2}

After receiving these instructions, participants were given two practice pairs that were manually generated to be obviously different and obviously similar and were required to rate these pairs above or below the midpoint of the scale, respectively, in order to progress to the main task.



\subsection{Subexperiment 3.1: Manually generated sentences}
\label{sec:subexperiment-3-1-manually-generated-sentences}

Subexperiment 3.1 has two purposes: (i) to assess the extent to which experts can reliably generate same-sense and different-sense pairs; and (ii) to produce a set of pairs of examples to be used in calibrating participants in distinctiveness experiments focused on evaluating our automated generation procedures.

\subsubsection{Materials}

As detailed in \Cref{sec:manually-generating-examples}, we constructed an extended calibration set for use in our distinctiveness experiments. To review, for each natural, typical sentence from our core calibration set---e.g. \Cref{ex:constant-sentence}---two additional sentences were generated: one that uses the same sense of the verb as the constant sentence and one that uses a different sense of the verb as the constant sentence.
Although not all of the calibration verbs have more than one PropBank sense, each pair of verbs in this set was ensured to differ in some way (based on our own judgment)---e.g. \Cref{ex:social-arrange} involves social and/or temporal arrangement, while \Cref{ex:physical-arrange} involves physical arrangement. 

\ex. 
\a. The baby arranged the blocks.\label{ex:constant-sentence}
\b. The florist arranged the flowers. \label{ex:physical-arrange}
\c. The secretary arranged the meeting.\label{ex:social-arrange}


\begin{table}[t]
\footnotesize
\centering
\caption{Examples of pairs of sentences used in the pair difference task.}
\label{tab:samediff}
\begin{tabular}{llll}
\toprule
    \textbf{Source} & \textbf{Sentences} & \textbf{Sense gloss} & \textbf{Pair type}  \\ \midrule
\multirow{4}{*}{Corpus} & The amateurs hit the road. & \texttt{hit.03} (go to, turn to) & \multirow{2}{*}{Same} \\
     & The resolutioners hit the gym. &  \texttt{hit.03} (go to, turn to)  &   \\ 
     \cmidrule{2-4}
& The amateurs hit the road. & \texttt{hit.03} (go to, turn to) & \multirow{2}{*}{Different} \\
   &  The body hit the floor. & \texttt{hit.02} (reach, encounter) \\
\midrule
\multirow{4}{*}{LM + PB senses} & The ball hit the wall.  & \texttt{hit.02} (reach, encounter) & \multirow{2}{*}{Same} \\
       & The ball hit the net. & \texttt{hit.02} (reach, encounter) &  \\ 
       \cmidrule{2-4}
   & The ball hit the wall.  & \texttt{hit.02} (reach, encounter) & \multirow{2}{*}{Different} \\
       & The road hit the lakefront. & \texttt{hit.03} (go to, turn to) &   \\ \midrule
\multirow{4}{*}{LM + LM senses} & The car hit the guardrail. & use violence or force & \multirow{2}{*}{Same} \\ 
                    & The police hit the rioters.  & use violence or force & \\ 
                    \cmidrule{2-4}
 & The car hit the guardrail.  & use violence or force & \multirow{2}{*}{Different} \\
        & The trumpeter hit the highnotes. & play a musical note \\ 
\bottomrule
\end{tabular}
\end{table}

\subsubsection{Participants}
We recruited 56 participants through Prolific to rate our manually-generated pairs, all who passed our practice questions. All participants were English speakers from the United States who had not participated in one of our previous studies. With IRB approval from our institutions, annotators were paid an average of \$12/hour.

\subsubsection{Selecting calibration sentences} 

The calibration pairs for the pair difference task were generated using the same procedure as those used in the naturalness and typicality tasks. However, the 30 pairs of calibration sentences were chosen such that all used unique verbs. Of the 30 pairs, 6 were shown to every single participant in Subexperiment 3.2 (described in the next subsection) after the examples and before any target sentences.

\subsection{Subexperiment 3.2: Automatically generated sentences}
\label{sec:subexperiment-3-2-automatically-generated-sentences}

\subsubsection{Materials}\label{sec:distinctiveness-materials}

For each way of automatically generating sentences, and for each sense of each verb---as labeled by a sense-tagger on the corpus; as elicited from the LLM using PropBank glosses; as elicited from the LLM using the LLM's own senses---we found the two sentences with the highest sum of their $z$-scored ratings for naturalness and typicality from Experiments 1 and 2. These two sentences represent a \textit{same-sense pair}.  Then, using the top sentence for each verb-sense---the one with the highest sum of $z$-scored naturalness and typicality---we created \textit{different-sense pairs}. So for a verb with $n$ senses, there are $n+\binom{n}{k}$ pairs.

\begin{table}[t]
\footnotesize
\centering
 \caption{Most and least different pairs of non-filler sentences by $z$-scored ratings, according to our participants. Notably, all \textit{most different} pairs contain different senses, while all \textit{least different} pairs contain sentences using the same sense of the verb.}
 \label{tab:pairdiffexample}
\begin{tabular}{lll}
\toprule
 & \textbf{Most different}  & \textbf{Least different}  \\
 \midrule
\multirow{ 2}{*}{Corpus} & The state passed the resolution. & The client mailed the letter. \\ 
& The grandmother passed the aunt. & The couple mailed the card.\\
\midrule
\multirow{ 2}{*}{LM + PropBank senses} & The actor belted the tune. & The soldier fired the rifle. \\ 
& The father belted the child. & The hunter fired the shotgun. \\
\midrule
\multirow{ 2}{*}{LM + LM senses} & The boy beat the drum. & The mother slapped the child. \\ 
& The team beat the rivals.  &  The parent slapped the teen. \\
\bottomrule
 \end{tabular}
\end{table}

Since this process yields more than 148 pairs of sentences per generation process, which is too many to be shown to a single participant, we implemented a similar list-making process to that described in \Cref{sec:subexperiment-1-2-automatically-generated-examples:materials}. Specifically, we randomly assigned each pair to a list number in ascending order such that no more than 60 unique pairs occur in the same list. Next, we padded each list with calibration sentences in their specified order until there were exactly 62 sentences per list. Since there are 2 practice questions and 6 sentences in the calibration block, this yielded lists with 70 sentences each. We wanted the distinctiveness surveys to have list rating instances than the naturalness and typicality surveys because in pilots, we found that distinctiveness ratings take participants more time.
In total, this process yields 3 lists generated for the survey with corpus-based sentences, 5 lists generated for the survey with LM sentences with Propbank senses, and 11 lists generated for the survey using LM sentences with LM senses.

\subsubsection{Participants}

We recruited 30 native English-speaking 
participants (of which all passed the practice questions) for the survey using LM sentences prompted with PropBank senses, 64 participants (of which 62 passed the practice questions) for the survey using LM sentences prompted with LM senses, and 20 participants (of which all passed the practice questions) for the survey using corpus-generated sentences on Prolific. No participants previously participated in any of our other surveys. Each participant was randomly shown a list from the list-making process described above.

\subsection{Results}


\begin{figure}[t]
  \centering
  \includegraphics[width=\textwidth]{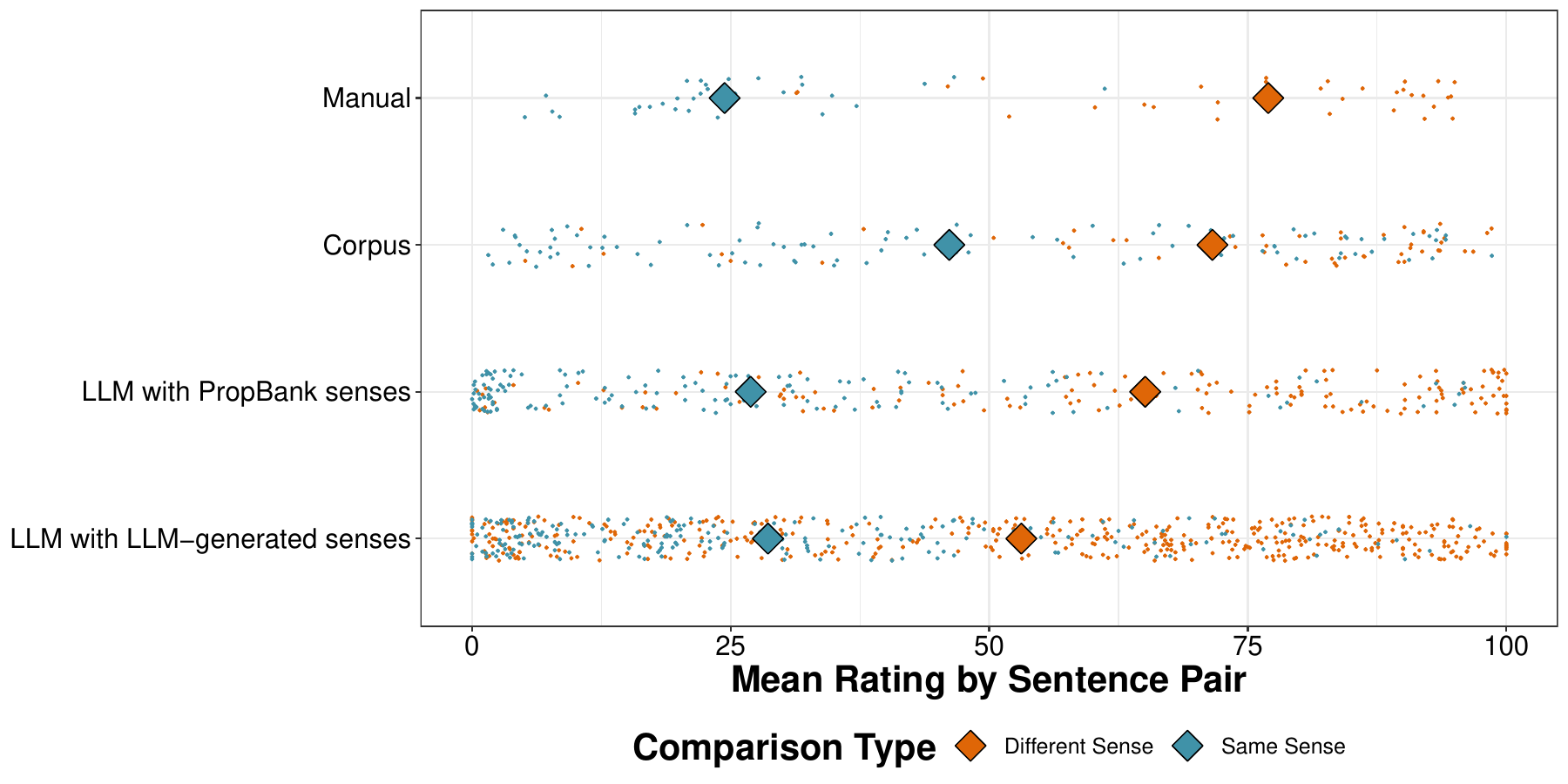}
  \caption{Mean difference rating for each sentence pair from each source. Each point shows the mean rating of a sentence pair and large colored points show the mean of those means for each relevant comparison type.}
  \label{fig:mean-diff}
\end{figure}

\Cref{fig:mean-diff} shows the mean distinctiveness ratings for each pair of sentences from each generation method, with examples of such pairs given in \Cref{tab:pairdiffexample}. Each black point shows the mean rating of a sentence and large colored points show the mean of those means. 
As expected, the manually generated pairs constructed to instantiate the same sense were rated more similar on average than those constructed to instantiate two different senses. 
All other generation methods show smaller differences between the same-sense pairs and the different-sense pairs.
In the case of the pairs constructed from corpus sentences, this difference appears to be due to the same-sense pairs being more different than the corresponding manual sense pairs.
In contrast, for the two LM-based methods, the different-sense pairs tend to be less different than the manually generated different-sense pairs.

\subsection{Analysis}

\begin{table}[t]
    \centering
    \small
    \caption{Fixed effect coefficient estimates in log-odds space for distinctiveness experiment with  \textit{Same Sense} as the reference level for comparison type and \textit{Manual} as the reference level for generation method in a dummy coding. The 2.5\% and 97.5\% columns give the lower and upper bound of the 95\% credible interval, respectively, and the posterior $p$ column gives the posterior probability that the coefficient has a sign different from the posterior mean. The posterior mean of the lower cutpoint of the ordered beta model is -2.51 (95\% CI = [-2.58, -2.43]) and the posterior mean of the mean of the upper cutpoints is 1.12 (95\% CI = [0.52, 1.83]).}
    \label{tab:difference-model-base}
\begin{tabular}{rrrrr}
\toprule
      & \textbf{Post. mean} & \textbf{2.5\%} & \textbf{97.5\%} & \textbf{Post.} $p$ \\
\midrule
Intercept & -1.48 & -1.89 & -1.08 & $<$ 0.01 \\
Different Sense & 1.91 & 1.40 & 2.43 & $<$ 0.01 \\
Corpus & 0.89 & 0.43 & 1.34 & $<$ 0.01 \\
LM with PropBank senses & -0.09 & -0.55 & 0.36 & 0.36 \\
LM with LM senses & 0.12 & -0.30 & 0.54 & 0.28 \\
Different Sense $\times$ Corpus & -1.14 & -1.82 & -0.47 & $<$ 0.01 \\
Different Sense $\times$ LM with PropBank senses & -0.29 & -0.90 & 0.33 & 0.18 \\
Different Sense $\times$ LM with LM senses & -0.87 & -1.44 & -0.30 & $<$ 0.01 \\
\bottomrule
\end{tabular}
\end{table}

To assess the reliability of the differences observed in \Cref{fig:mean-diff}, we fit an ordered beta mixed effects model to the responses. 
This model had fixed effects for generation method---treating classes of manually generated sentences as separate generation methods---as well as by-participant, by-verb, by-sense pair, and by-sentence pair random intercepts. \Cref{tab:difference-model-base} shows the estimates for the fixed effect coefficients. 

Consistent with the qualitative pattern observed in \Cref{fig:mean-diff}, the different-sense pairs are rated as reliably more different than the same-sense pairs for the manual sentences.
Also consistent with the qualitative pattern observed in \Cref{fig:mean-diff}, the same-sense corpus sentences are rated as reliably more different than the manually generated same-sense pairs.
The interaction of corpus-sampling with different-sense pairs furthermore has a high posterior probability of being negative.
This interaction term is slightly larger in magnitude that the simple effect of being generating based on a corpus sampling, capturing the pattern observed in \Cref{fig:mean-diff}: the different-sense pairs that come from the corpus appear to be rated as slightly less different on average than the manually generated different sense pairs.
This difference is not reliable, however;
the posterior probability that the manually generated different-sense pairs are rated as more different than the corpus-sampled different-sense pairs is $\sim$0.80. 

Also consistent with the qualitative pattern observed in \Cref{fig:mean-diff}, the same-sense pairs for both LM-based methods are not rated reliably more or less different than the manually generated pairs.
The two methods split apart in how their different-sense pairs behave: when conditioning on LM-generated senses, the resulting pairs are reliably less different than under manual generation;
but when conditioning on manually generated (PropBank) sense glosses, the difference is smaller and not reliable.

When comparing the to LM-based methods to each other, the different-sense pairs when conditioning on manually generated senses are rated as reliably more different than when conditioning on LM-generated sense glosses (posterior $p <$ 0.01).
This pattern suggests that conditioning sentence generation on manually generated sense glosses when generating from an LM yields a more distinctive set of sentences than conditioning on LM-generated sense glosses.

Further, while different-sense pairs that come from the corpus appear in \Cref{fig:mean-diff} to be rated as more different than the LM-generated pairs that use manually generated senses, this apparent difference is not reliable (posterior $p =$ 0.27).
Indeed, neither is the difference between manually generated different-sense pairs and LM-generated pairs that use manually generated senses (posterior $p =$ 0.08)
This result is likely a product of the high variability in the disitnctiveness ratings for the LM-generated different-sense pairs.

\subsection{Discussion}

We find that event descriptions produced by the automatic methods we consider are not as distinctive as our best manually generated sentences but that they are nonetheless substantially more distinctive than event descriptions manually constructed to use the same sense. 
We conclude from this result that all of the automated generation methods we consider are safe to use for generating sentences for downstream annotation insofar as some small amount of degradation in distinctiveness is tolerable.

Which automated method is preferable in any particular context depends on the structure of the downstream experiment.
For instance, if one is purely concerned with generating a distinctive set of sentences, generating sentences either on the basis of corpus samples or from an LM-conditioned a manually generated sense lexicon will likely yield negligibly different results---though the presence of high variability for the LM-based generation may suggest a larger sample is necessary to ensure that the average distinctiveness is sufficiently high.
Whether such a larger sample can be feasibly annotated depends on the size of the class of verbs of interest and the range of structures that satisfy the syntactic constraints of interest.
If, on the other hand, one is particularly concerned with ensuring a particular proportion of similar and distinctive items, LM-generation conditioned on manually generated senses may be the best choice. 
 
\section{Conclusion} 
\label{sec:conclusion}

In three experiments, we have demonstrated that automated methods for generating linguistic expressions for downstream annotation and analysis can generate reliably natural, typical, and distinctive event descriptions across different senses of verbs---though all automated methods we investigated yield less natural, typical, and distinctive linguistic expressions than manual generation can---summarized in \Cref{tab:final-summary}.
We draw two main conclusions from these findings. 
\begin{table}[t]
\centering
\caption{Final summary of sampling methods for linguistic stimuli (i.e., sentences).}
\label{tab:final-summary}
\begin{tabular}{rccc}
\toprule
                & \textbf{Quality} & \textbf{Effort} & \textbf{Efficiency} \\
                \midrule
\textbf{Manual} & High             & High             & Low                 \\
\textbf{Corpus} & \fbox{Medium}                & Low              & Medium              \\
\textbf{LLM}    & \fbox{Medium}               & Low              & High                \\
\bottomrule
\end{tabular}

\end{table}
First, we conclude that, while the automated methods can be used to generate linguistic expressions of reasonably good quality, they should largely be reserved for approaches to data collection and analysis (i) in which it is simply not feasible to manually generate all linguistic expressions of interest; and (ii) that can be made robust to degradation in linguistic expressions' naturalness, typicality, and distinctiveness.
Effectively, we submit that automated generation methods are a good option for lexical semantic research wherein large portion of the lexicon are being studied---e.g. if one were interested in drawing sweeping generalizations about all clause-embedding predicates and were worried about known methodological issues that are known to arise from poor sampling methodologies \citep[see][]{white_believing_2021}---but that there is little reason to use them if smaller portions of the lexicon are under investigation. 

Second, we conclude that LM-based methods yield linguistic expressions of sufficiently comparable quality to those yielded by corpus-based methods---at least under the sorts of heavy constraints we impose in this paper---that they are preferable (all else being equal) to corpus-based methods for their efficiency.

These results raise a number of potentially interesting questions for future research.
First, how well do the automated methods we consider---and the LM-based methods in particular---generalize to more complex syntactic constraints? We considered relatively strict constraints, resulting in a relatively simple syntactic context for the verbs of interest. Do the automated methods we consider gracefully scale to constraints that result in more complex syntactic structures?

Second, how well do the automated methods we consider generalize to more complex semantic or pragmatic constraints? For instance, if we are interested in generating multisentence contexts, how far can a combination of prompting and constrained decoding from an LM take us in the absence of nontrivial modifications to that LM?

Finally, to what extent can we improve the outputs of the automated methods we consider in an efficient way? For instance, is it possible to deploy efficient post-editing of automatically generated linguistic expressions (as in, e.g., \citealt{green_efficacy_2013}, i.a.) to bring the quality of those expressions to the level of fully manually generated expressions in a way that reduces the overall human effort required by full manual generation?

\section{Acknowledgments}

We gratefully acknowledge the support of National Science Foundation collaborative grant BCS-2040831/2040820 (\textit{Computational Modeling of the Internal Structure of Events}).

\starttwocolumn
\bibliography{references}

\end{document}